\documentclass[runningheads]{llncs}

\usepackage{amssymb}
\usepackage{amsmath}
\usepackage{graphicx}
\usepackage[ruled,vlined]{algorithm2e}
\usepackage{algorithmic}
\usepackage{multirow}
\usepackage{array}
\usepackage{float}
\usepackage{caption}

\usepackage{color,soul} 
\makeatletter
 \def\SOUL@hlpreamble{%
 \setul{}{2.4ex}
 \let\SOUL@stcolor\SOUL@hlcolor
 \SOUL@stpreamble
 }
\makeatother

\begin{document}

\mainmatter  

\title{Quantification of Local Metabolic Tumor Volume Changes by Registering Blended PET-CT Images for Prediction of Pathologic Tumor Response}

\titlerunning{Blended PET-CT}  

\author{Sadegh Riyahi\inst{1} \and Wookjin Choi\inst{1} \and Chia-Ju Liu\inst{1} \and Saad Nadeem\inst{1} \and Shan Tan\inst{2} \and Hualiang Zhong\inst{3} \and Wengen Chen\inst{4} \and Abraham J. Wu\inst{1} \and James G. Mechalakos\inst{1} \and Joseph O. Deasy\inst{1} \and Wei Lu\inst{1}}

\authorrunning{Riyahi et al.} 

\tocauthor{Sadegh Riyahi, Wookjin Choi, Chia-Ju Liu, Saad Nadeem, Shan Tan, Hualiang Zhong, Wengen Chen, Abraham J. Wu, James G. Mechalakos, Joseph O. Deasy, and Wei Lu}

\institute{Memorial Sloan Kettering Cancer Center, New York, USA
\and
Huazhong University of Science and Technology, Wuhan, China
\and
Henry Ford Hospital, Detroit, USA
\and
University of Maryland School of Medicine, Baltimore, USA}

\maketitle              

\begin{abstract}
Quantification of local metabolic tumor volume (MTV) chan-ges after Chemo-radiotherapy would allow accurate tumor response evaluation. Currently, local MTV changes in esophageal (soft-tissue) cancer are measured by registering follow-up PET to baseline PET using the same transformation obtained by deformable registration of follow-up CT to baseline CT. Such approach is suboptimal because PET and CT capture fundamentally different properties (metabolic vs. anatomy) of a tumor. In this work we combined PET and CT images into a single blended PET-CT image and registered follow-up blended PET-CT image to baseline blended PET-CT image. B-spline regularized diffeomorphic registration was used to characterize the large MTV shrinkage. Jacobian of the resulting transformation was computed to measure the local MTV changes. Radiomic features (intensity and texture) were then extracted from the Jacobian map to predict pathologic tumor response. Local MTV changes calculated using blended PET-CT registration achieved the highest correlation with ground truth segmentation (R=0.88) compared to PET-PET (R=0.80) and CT-CT (R=0.67) registrations. Moreover, using blended PET-CT registration, the multivariate prediction model achieved the highest accuracy with only one Jacobian co-occurrence texture feature (accuracy=82.3\%). This novel framework can replace the conventional approach that applies CT-CT transformation to the PET data for longitudinal evaluation of tumor response.
\end{abstract}

\section{Introduction}
Image-based quantification of tumor change after Chemo-radiotherapy (CRT) is important for evaluating treatment response and patient follow-up. Standard methods to assess the tumor metabolic response in Positron Emission Tomography (PET) images are qualitative and described based on a discrete categorization of reduction in Standardized Uptake Value (SUV) or Metabolic Tumor Volume (MTV) \cite{westerterp2005esophageal}. Overall volumetric difference is a global measurement that cannot characterize local non-uniform changes after the therapy \cite{westerterp2005esophageal}. For these reasons, diameter/SUV/volume based measurements are not consistently correlated to important outcomes \cite{westerterp2005esophageal}. Tensor Based Morphometry \cite{riyahi2018jac} exploits the gradient of Deformation Vector Field (DVF) i.e. determinant of Jacobian matrix termed Jacobian map (J), to characterize voxel-by-voxel volumetric ratio of an object before and after the transformation. J $>$ 1 means local volume expansion, J $<$ 1 means shrinkage and J = 1 denotes no change. There are many studies that utilize Jacobian map to evaluate volumetric changes. Fuentes et al. \cite{fuentes2015morphometry} used Jacobian integral (mean J$\times$tumor volume) to measure the local volume change of irradiated whole-brain tissues in Magnetic Resonance Images and showed that the estimated change had good agreement with ground truth segmentation. In our previous work \cite{riyahi2018jac} we showed that Jacobian features in Computed Tomography (CT) images could predict the tumor pathologic response with high accuracy (94\%) in esophageal cancer patients.

However, structural change in CT is affected by daily anatomical variations and therapy response is mostly seen in PET as metabolic activity \cite{westerterp2005esophageal}. Conventionally, metabolic tumor change is measured by deforming the follow-up tumor volume in PET and aligning it to baseline tumor volume using the transformation obtained from CT-CT Deformable Image Registration (DIR) \cite{van2014effects}. However, PET and CT capture different properties (metabolic vs. anatomy) of a tumor, therefore applying the transformation from CT-CT registration is suboptimal. On the other hand, directly registering PET images is problematic since there are few image features to generate an accurate transformation \cite{van2014effects}.

Some attempts performed on deformable registration of PET-CT using joint maximization of intensities \cite{Jin2013petctreg} increased the uncertainties due to heterogeneous tumor uptake in PET and different intensity distributions between two images. Additionally, deep learning methods to estimate DVF have been proposed recently. However, training deformations were generated using existing Free-Form registrations, hence the accuracy could be as good as already available algorithms \cite{krebs2017deep}. Moreover, the algorithms were not tested for multi-modality registrations.

In this work, we used a linear combination of PET and CT images to generate a single grayscale blended PET-CT image using a pixel-level fusion method. Our main goal is to combine anatomic and metabolic information to improve the accuracy of multi-modality PET-CT registration for quantification of tumor change and for prediction of pathologic tumor response. The contributions are as follows:
\begin{enumerate}
    \item Local MTV change calculated using Jacobian integral of blended PET-CT image registration achieved higher correlation with the ground truth segmentation (R=0.88) compared to mono-modality PET-PET (R=0.80) and CT-CT (R=0.67) registrations.
    \item Jacobian radiomic features extracted from blended PET-CT registration could better differentiate pathologic tumor response (AUC=0.85) than mono-modality PET and CT Jacobian and clinical features (AUC=0.65$\sim$0.81) with only one Jacobian co-occurrence texture feature in esophageal cancer patients. 
\end{enumerate}

\section{Material and Methods}

\begin{figure*}[t!]
\begin{center}
\begin{tabular}{cccc}
\includegraphics[width=0.6\textwidth]{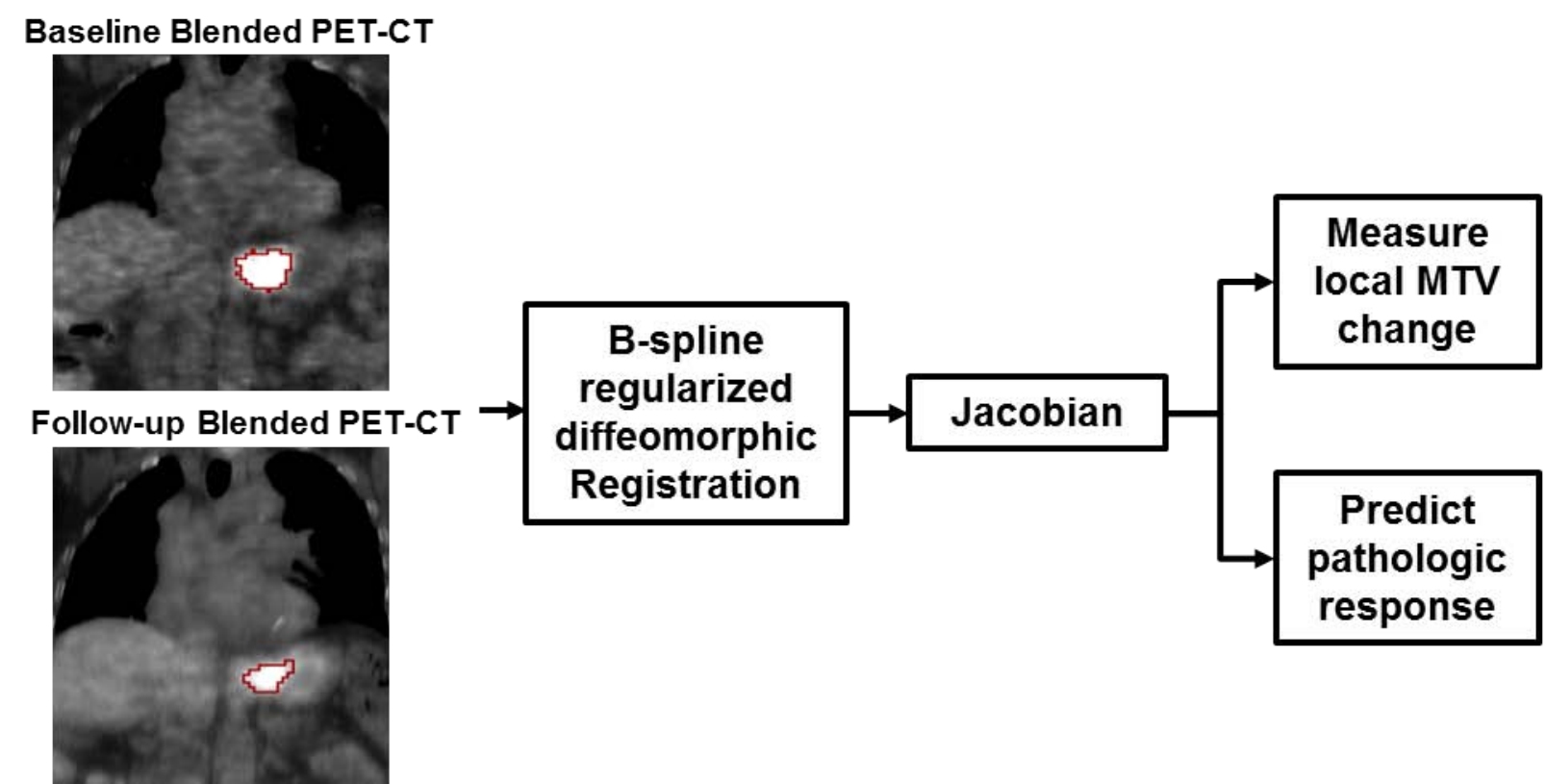}&
\includegraphics[width=0.17\textwidth]{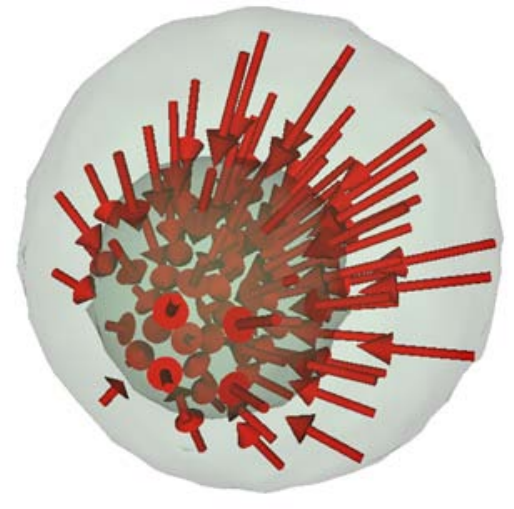}&
\includegraphics[width=0.17\textwidth]{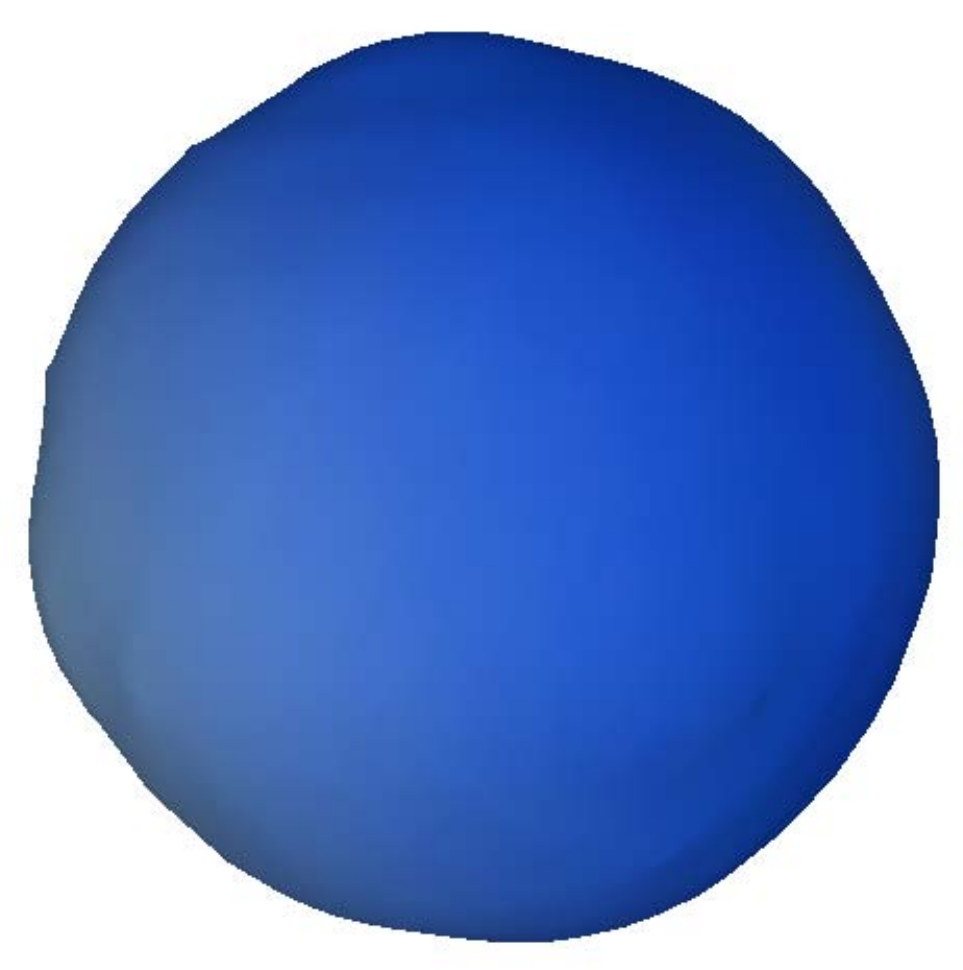}&
\includegraphics[width=0.035\textwidth]{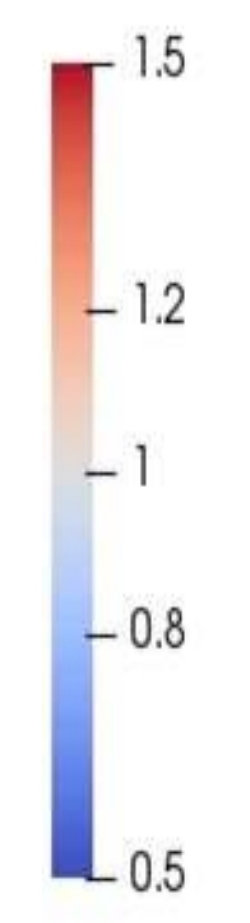}\\
(a) & (b) & (c)
\end{tabular}
\end{center}
\caption{(a)Main workflow of our method. Conceptual illustration of Jacobian map. (b)Larger sphere simulates MTV in the baseline image and smaller follow-up sphere illustrates shrinkage of a tumor. Converging DVF represents a volume loss and generates a Jacobian map (c) that illustrates local shrinkage (blue).
\label{fig:pipeline}}
\end{figure*}


Fig. \ref{fig:pipeline} shows our workflow and illustrates the concept of Jacobian map using a synthetic sphere that simulates a heterogeneous tumor shrinkage. 

\subsection{Dataset}

This study included 61 patients with esophageal cancer who were treated with induction chemotherapy followed by CRT and surgery. All patients underwent baseline, post-induction and post-CRT PET/CT scans. Resolution for PET images was $4.0\times4.0\times4.25$ $mm^3$ and for CT images was $0.98\times0.98\times4.0$ $mm^3$. MTV on each PET-CT was segmented using a semi-automatic adaptive region-growing algorithm developed by our group \cite{tan2017adaptive}. Segmentations were visually reviewed and manually modified if necessary by a nuclear medicine physician. Average percentage of MTV change was 50$\pm$30.6\% in the cohort. Pathologic tumor response was assessed in surgical specimen and categorized into: pathologic complete responders (absence of viable tumor cells, 6 patients) and non-responders (partial response, progressive or stable disease, 55 patients). Registrations were performed between baseline and post-induction chemotherapy (follow-up) images.

\subsection{Generating blended PET-CT images}
Maximum intensity of CT images was clipped to 750 HU to eliminate the effect of high attenuation metals. PET images were resampled to CT resolution. PET and CT images were normalized to the range of [0, 1]. The normalization bounds used for CT were (-1000, 750) HU and for PET, the range of tumor SUVs in our patient cohort (0, 35) was used. To generate a grayscale blended PET-CT image, a weighted sum of normalized PET (nPET) and CT images (nCT) was formulated (Eq. 1) where $\alpha \in [0,1]$:
\begin{equation}
    Blended\; PETCT = \alpha (nCT) + (1-\alpha) nPET    
\end{equation}
\( \alpha\)=0.2 was found optimal in that it produced similar blending of PET and CT information as when the nuclear medicine physician visually fused PET (window/level = 6/3 SUV) and CT (window/level = 350/40 HU) images. By using blended PET-CT images for registration, high metabolic uptake in the tumor was emphasized in the foreground while anatomic details in surrounding normal tissues were kept in the background (Fig. \ref{fig:main_fig}).

\subsection{Registration Methods}
\emph{\bf B-spline regularized diffeomorphic registration:} To correct respiratory-induced tumor motion, we first aligned follow-up images to baseline images by rigidly registering the tumors using their center of geometry as an initial transformation. Then we deformably registered two images using a rigidity penalty term \cite{staring2007rigidity} to enforce the local rigidity on tumor and preserve tumor's structure while compromising on the global surrounding differences. Rigidity penalty was only applied to blended PET-CT and PET-PET registrations. Initial alignment of CT images was performed using a rigid registration. We then deployed a B-spline regularized symmetric diffeomorphic registration (BSD) \cite{tustison2013explicit} to characterize metabolic volume loss. A diffeomorphic registration estimates the optimized transformation, $\phi$, parameterized over $t \in [0,1]$ that maps the corresponding points between two images. $\phi$ is obtained by a Symmetrized Large Deformation Diffeomorphic Metric Mapping (LDDMM) algorithm that finds a geodesic solution in the space of diffeomorphism. A symmetrized LDDMM captures large intra-modality differences and guarantees inverse consistency and one-one mapping in DVF while minimizing the bias between forward and inverse transformations. By explicitly integrating the B-spline regularization term, a viscous-fluid model is formulated that fits the calculated DVF after each iteration to a B-spline object. This gives free-form elasticity to converging vectors creating a sink point that is mapped to many points in its vicinity and represents a morphological shrinkage for the regions with non-mass conserving deformations. The optimization cost function is as follows \cite{tustison2013explicit}:
\begin{equation}
    \begin{array}{r}
         c(\phi(x,t),I_b\circ I_f )= E_{similarity}^{MI} (\phi(x,1),I_b,I_f) +
         E_{geodesic}^2 (\phi(x,0),\phi(x,1)) \\ + \rho_{Bspline} (v(\phi(x,t)),B_k)
    \end{array}
\end{equation}
where $E_{similarity}^{MI}$ is a mutual information similarity energy, $E_{geodesic}$ is a geodesic energy function and $\rho_{Bspline}$ denotes a B-spline regularizer. The transformation $\phi(x,t)$ between baseline ($I_b$) and follow-up ($I_f$) images is characterized by the maps of the shortest path between time points $t=0$ and $t=1$.
\newline
\noindent
$v(\phi(x,t)) = \frac{\partial\phi(x,t)}{\partial t}$ is the gradient field that defines the displacement change at any given time point. $B_k$ is B-spline function (k spline order) applied on the gradient field. Three levels of multi-resolution registration were implemented with B-spline mesh size of 32mm, 32mm and 16mm at the coarsest level for blended PET-CT, PET-PET and CT-CT registrations, respectively. The mesh size was reduced by a factor of 2 at each sequential level. The optimization step size was set to 0.15 and the number of iterations (100,70,40) for all modalities. We used Directly Modified Free Form Deformation optimization scheme \cite{tustison2013explicit} that was robust to different parameters and all the registrations were performed in a cropped region 5cm surrounding the MTV.
\newline
\newline
\noindent
\emph{\bf Registration evaluation methods:} We considered MTV change measured by the semi-automatic segmentation with physician modification as the ground truth to compare against Jacobian integral for registration evaluation. Correlation and percentage of difference between MTV changes calculated by Jacobian integral and by semi-automatic segmentation (ground-truth) were first assessed. Dice Similarity Coefficient (DSC) was also calculated between baseline MTV and deformed follow-up MTV. We compared BSD results with a Free-Form Deformation Registration algorithm (FFD) regularized with bending energy \cite{klein2010elastix}. The blended PET-CT, PET-PET and CT-CT registrations were separately performed using these two algorithms.
\newline
\newline
\noindent
\emph{\bf Optimal registration parameter estimation:} i) Regularization mesh size ($\sigma$) and ii) optimization step size ($\gamma$) were the most sensitive parameters. We experimentally studied the influence of different $\sigma$ = 16, 32, 64, 128 mm and \newline $\gamma$ = 0.1, 0.15, 0.2, 0.25 on registration and Jacobian map. The registration results were used as a quantitative benchmark to find the optimal trade-off between the parameters. A parameter set that resulted in the best DSC and the highest correlation between Jacobian integral and segmentation was chosen as the optimal parameters.
\subsection{Jacobian Features for Prediction of Tumor Response}
We extracted 56 radiomic features quantifying the intensity and texture \cite{yip2016textureregistration} of a tumor in the Jacobian map. The Jacobian features quantified the spatial patterns of tumor volumetric change. The importance of features in predicting pathologic tumor response was evaluated by both univariate and multivariate analysis. In univariate analysis, p-value and Area Under the Receiver Operating Characteristic Curve (AUC) for each feature was calculated using Wilcoxon rank sum test.
In multi-variate analysis, firstly distinctive features were identified using hierarchical clustering \cite{choi2018medphy}.
A Random Forest model (RF) was then constructed (200 trees) with features chosen by a Least Absolute Shrinkage and Selection Operator (LASSO) feature selection. All distinctive features were fed to the RF classifier in a manner of a 10-fold cross-validation (CV). Within each fold, LASSO was applied to select the ten most important features. We repeated the 10-fold CV ten times to obtain the model accuracy (10x10-fold CV).

\section{Results and Discussion}

\subsection{Quantitative registration evaluation} 
\begin{figure*}[htb]
\begin{center}
\begin{tabular}{ccc}
\includegraphics[width=0.353\textwidth,height=0.353\textwidth]{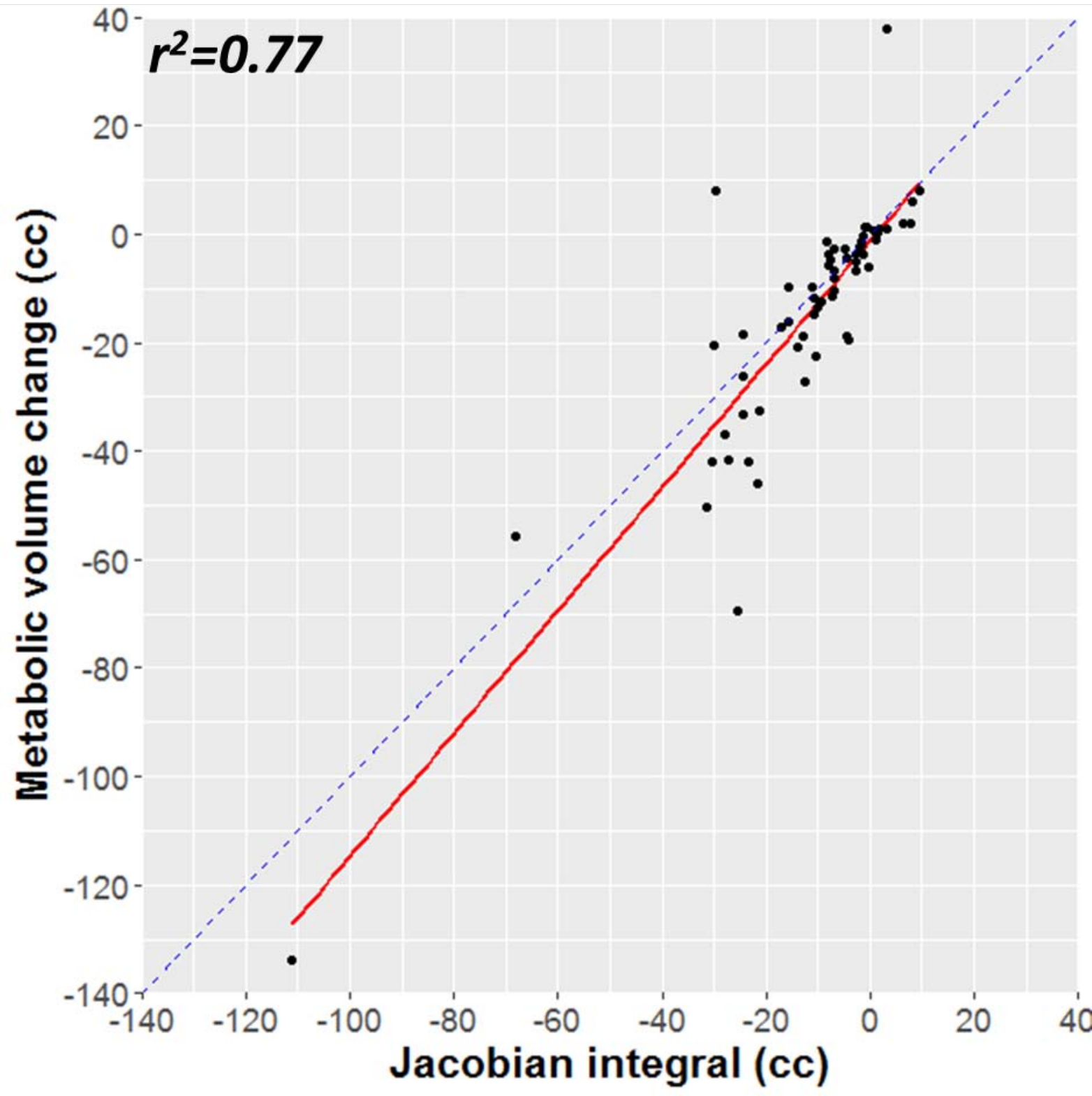}&
\includegraphics[width=0.315\textwidth,height=0.353\textwidth]{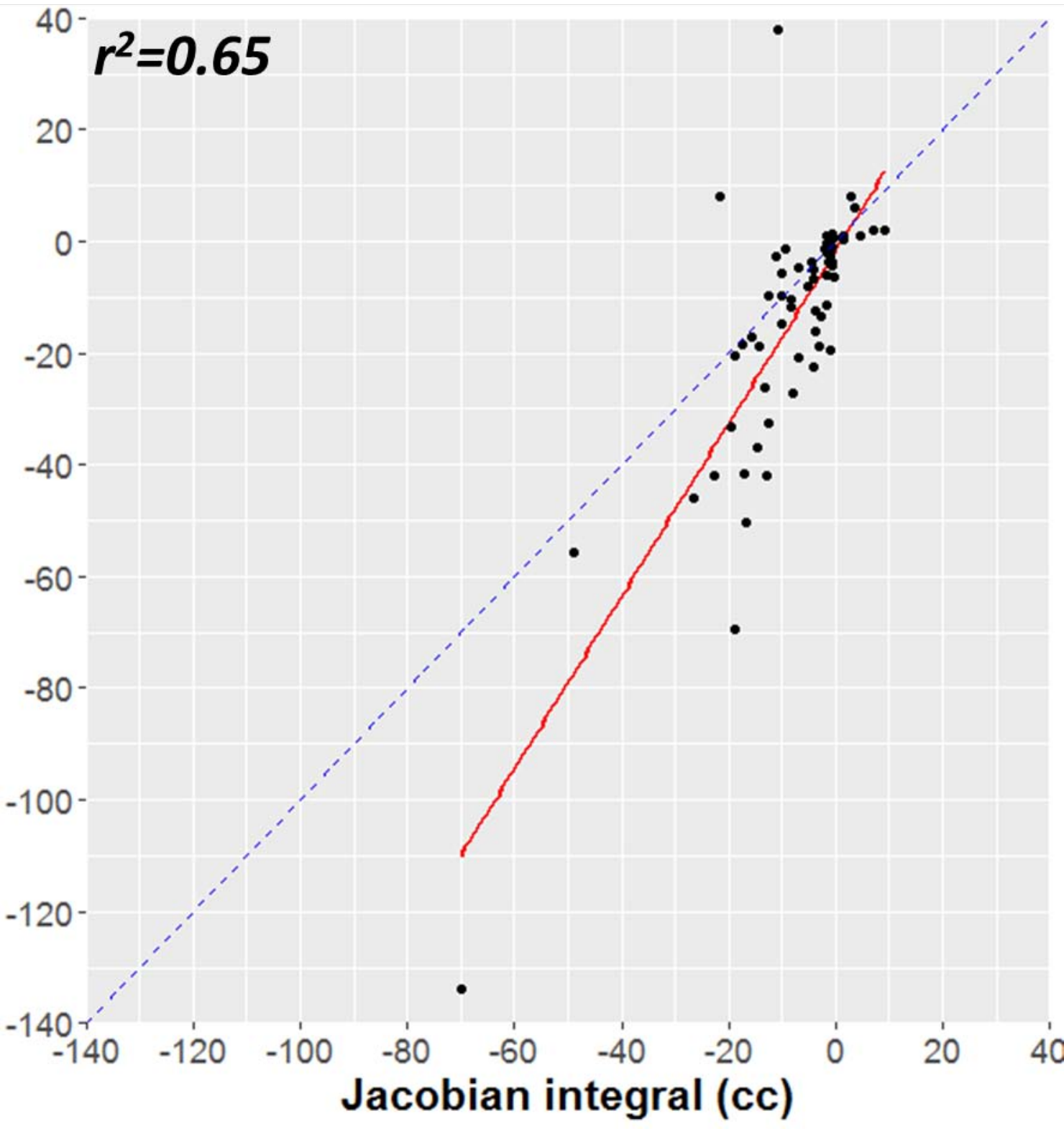}&
\includegraphics[width=0.315\textwidth,height=0.353\textwidth]{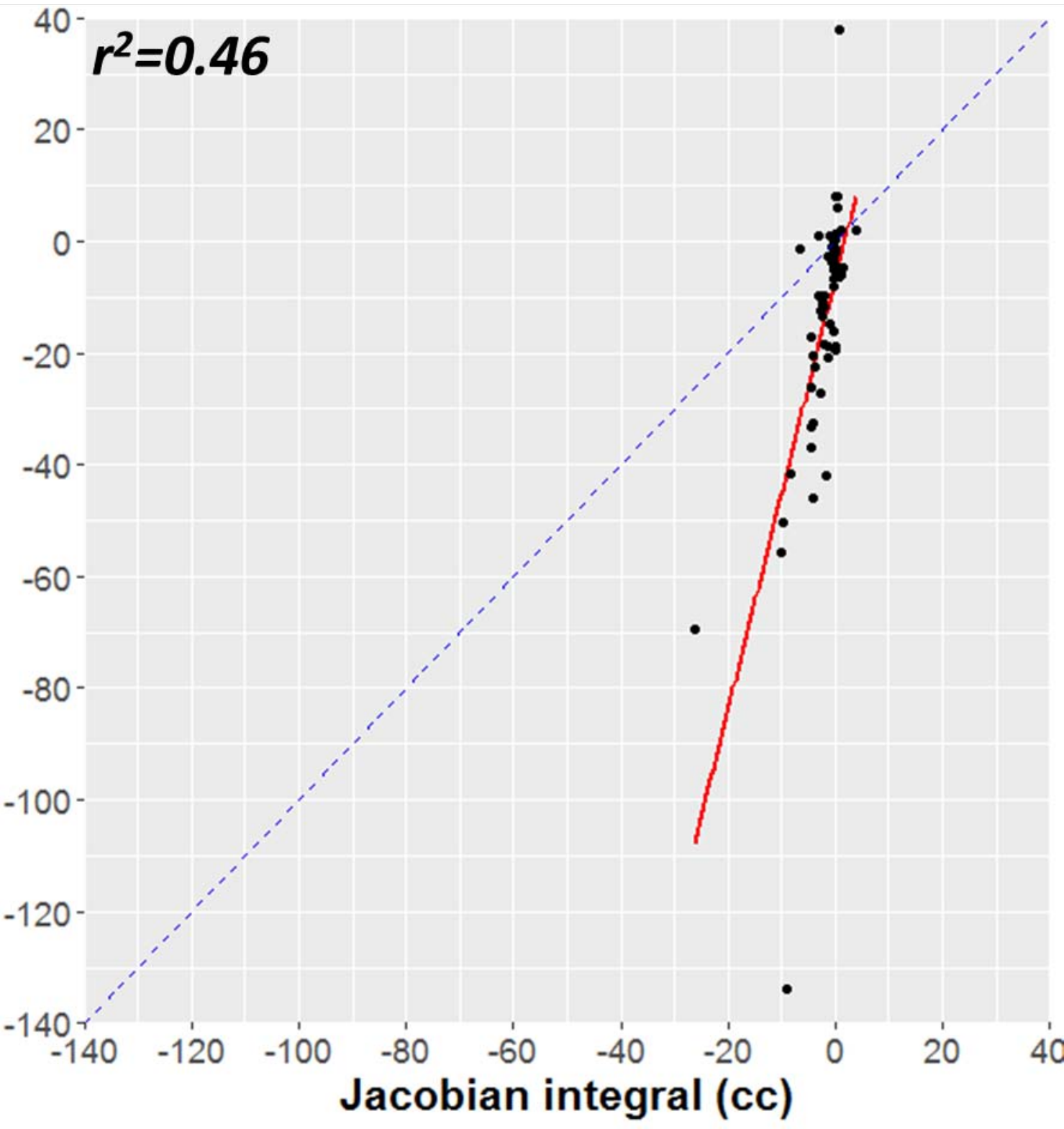}\\
(a) Blended PET-CT & (b) PET-PET & (c) CT-CT
\end{tabular}
\end{center}
\caption{Scatter plot showing correlation between MTV change calculated by Jacobian integral and ground truth segmentation for (a) blended PET-CT, (b) PET-PET and (c) CT-CT BSD registrations. Dashed blue line is identity line.
\label{fig:scatter}}
\end{figure*}

A combination of $\sigma$=32mm (blended PET-CT), 32mm (PET-PET), 16mm (CT-CT) and $\gamma$=0.15 achieved the best DSC and the highest correlation hence were selected as the optimal registration parameters. Larger mesh size in blended PET-CT and PET-PET registrations compared to CT-CT registration produced a more regularized and smoothed DVF to compensate the local irregular deformations due to non-uniform metabolic uptakes and lack of corresponding points in PET. Fig. \ref{fig:scatter} shows scatter plots with least square regression line (solid red) between MTV change calculated by Jacobian integral and the ground truth segmentation for (a) blended PET-CT, (b) PET-PET and (c) CT-CT BSD registrations with goodness of fit ($r^2$) values. Blended PET-CT registration showed the highest $r^2$ and captured the greatest range of deformations in tumor, compared to PET-PET and CT-CT registrations. Table \ref{table:registration_eval} shows correlation coefficients and average percentage of difference between Jacobian integral and segmentation using BSD and FFD for each modality.

Mean$\pm$stdev DSC are also presented in Table \ref{table:registration_eval}. Blended PET-CT registration showed higher DSC with less variation among the cohort. Using a blended PET-CT registration, DVF in and near the tumor region was driven by the metabolic changes while DVF outside the tumor region was driven by the anatomical structures surrounding the tumor. The blended PET-CT registration benefited by leveraging prominent image features from both PET and CT simultaneously, hence, achieving higher DSC and more accurate estimation of MTV change.
\newline
\begin{table}[t]
\centering
\caption{Registration results using the optimal parameters comparing correlation and average percentage difference between MTV change estimated by Jacobian integral and segmentation.}
\label{table:registration_eval}
\setlength{\tabcolsep}{4pt}
\begin{tabular}{c|c|c|c|c}
\hline
\bf Registration & \bf Correlation & \bf \% Difference & \bf DSC & \bf Quantified changes\\
\hline
Segmentation & - & - & - & 50\%\\
\hline
PET-CT BSD & 0.88 & 7.8\% & 0.73$\pm$0.08 & 42\%\\
PET BSD	   & 0.80 & 14.1\% & 0.66$\pm$0.13 & 28.6\%\\
CT BSD	   & 0.67 & 31.6\%	& 0.69$\pm$0.16 & 7.6\%\\
\hline
PET-CT FFD & 0.77 & 18.6\% & 0.71$\pm$0.13 & 22\%\\
PET FFD    & 0.74 &	25.1\% & 0.60$\pm$0.14 & 17.4\%\\
CT FFD	   & 0.32 & 28.4\% & 0.69$\pm$0.16 & 19.6\%\\
\hline
\end{tabular}
\end{table}

\subsection{Residual tumor versus non-residual tumor cases} 
\begin{figure*}[t!]
\begin{center}
\begin{tabular}{cc@{\hskip 0.08in}|@{\hskip 0.08in}c@{\hskip 0.08in}|@{\hskip 0.08in}c@{\hskip 0.08in}|@{\hskip 0.08in}cc}
\footnotesize Baseline & \footnotesize Follow-up & \footnotesize Blended PET-CT & \footnotesize PET-PET & \footnotesize CT-CT & \\
\includegraphics[width=0.13\textwidth]{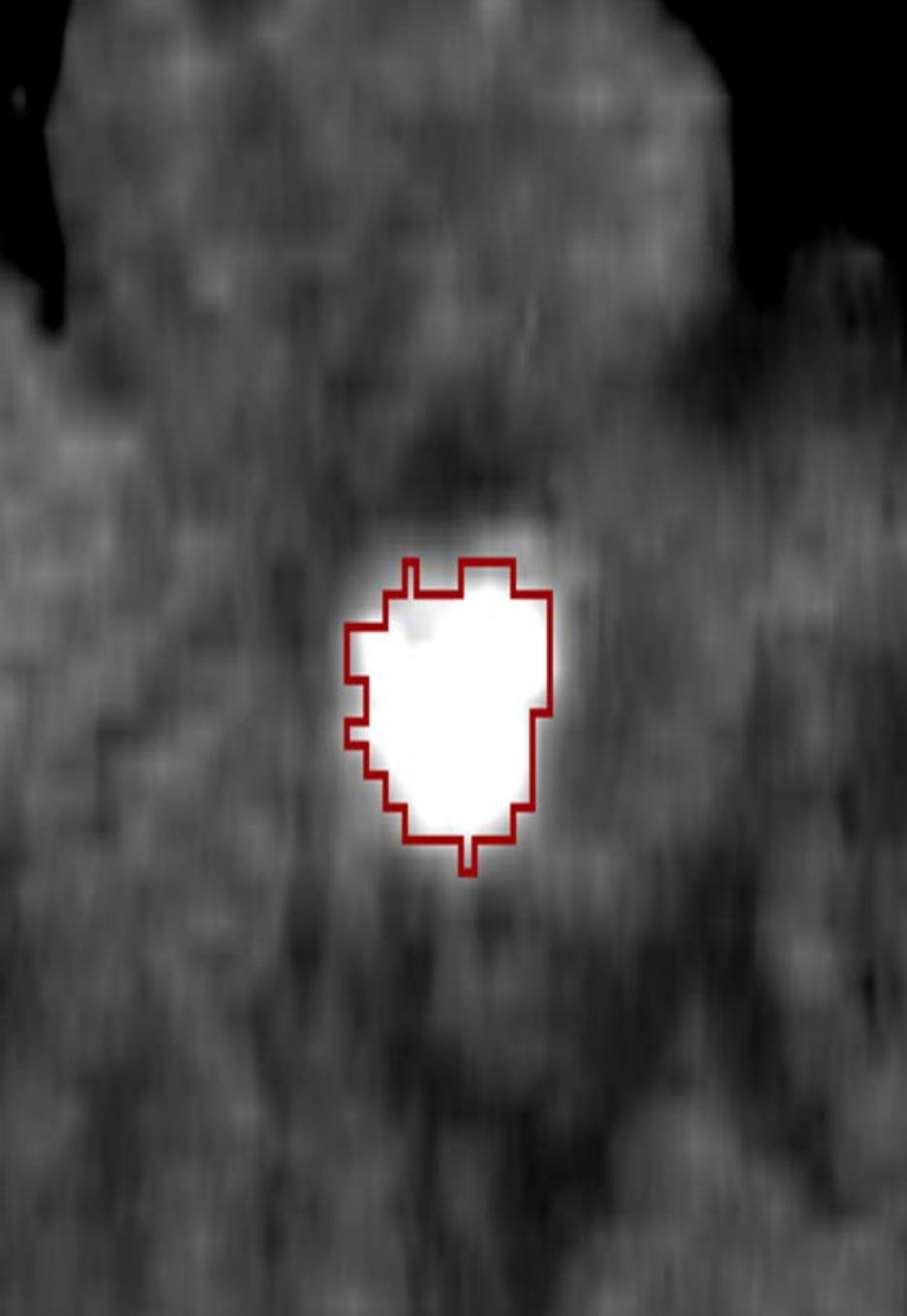}&
\includegraphics[width=0.13\textwidth]{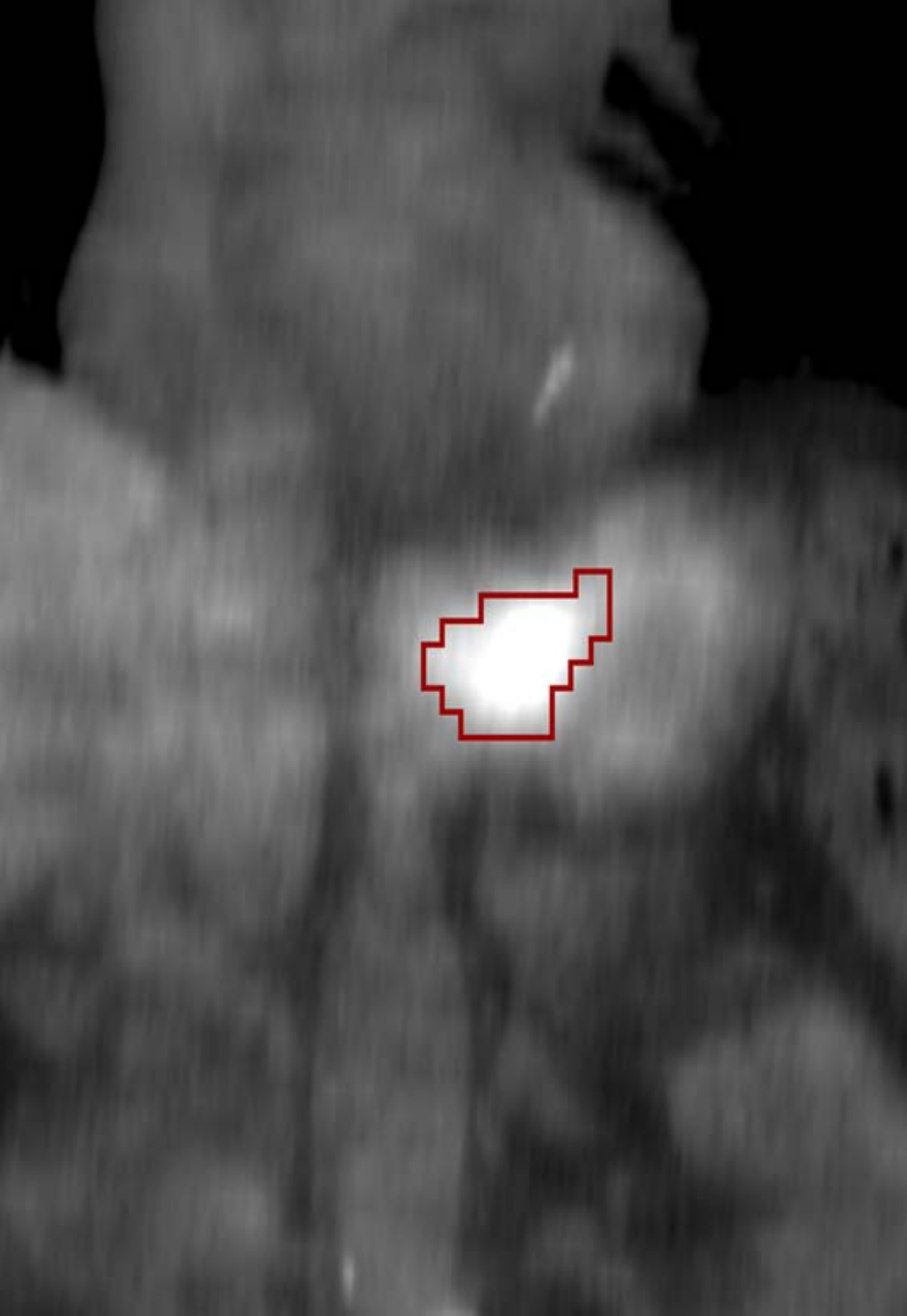}&
\includegraphics[width=0.20\textwidth]{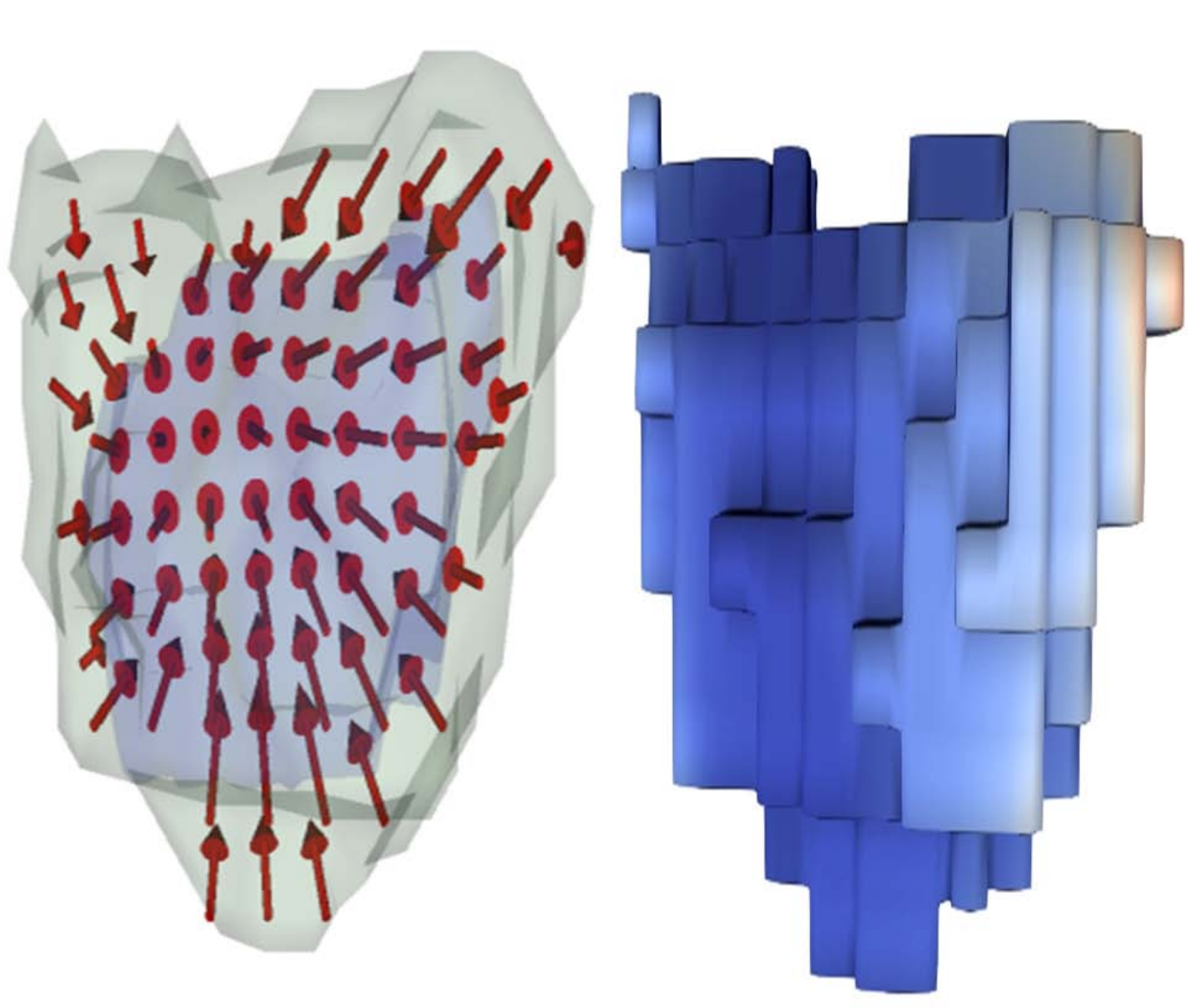}&
\includegraphics[width=0.20\textwidth]{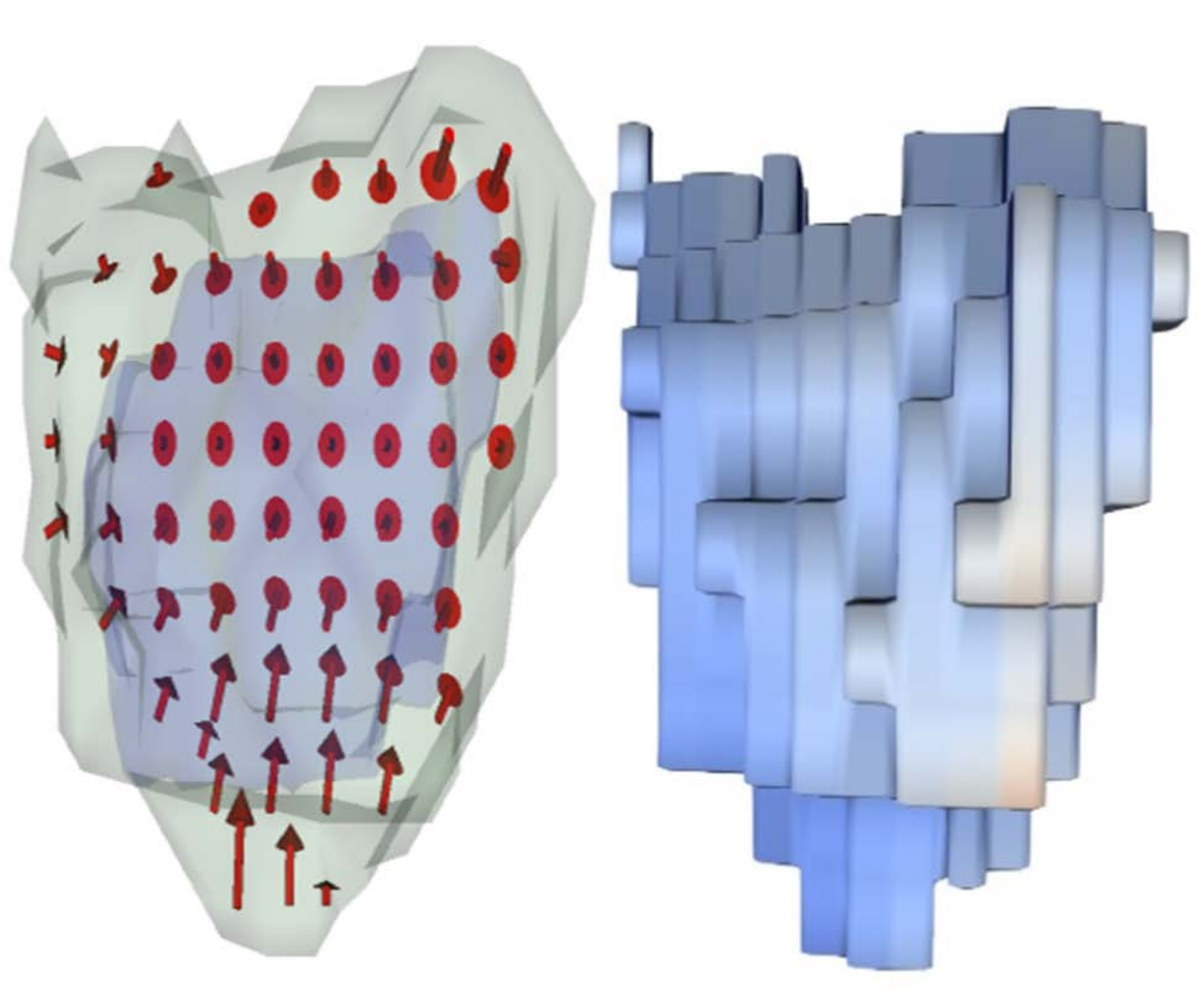}&
\includegraphics[width=0.20\textwidth]{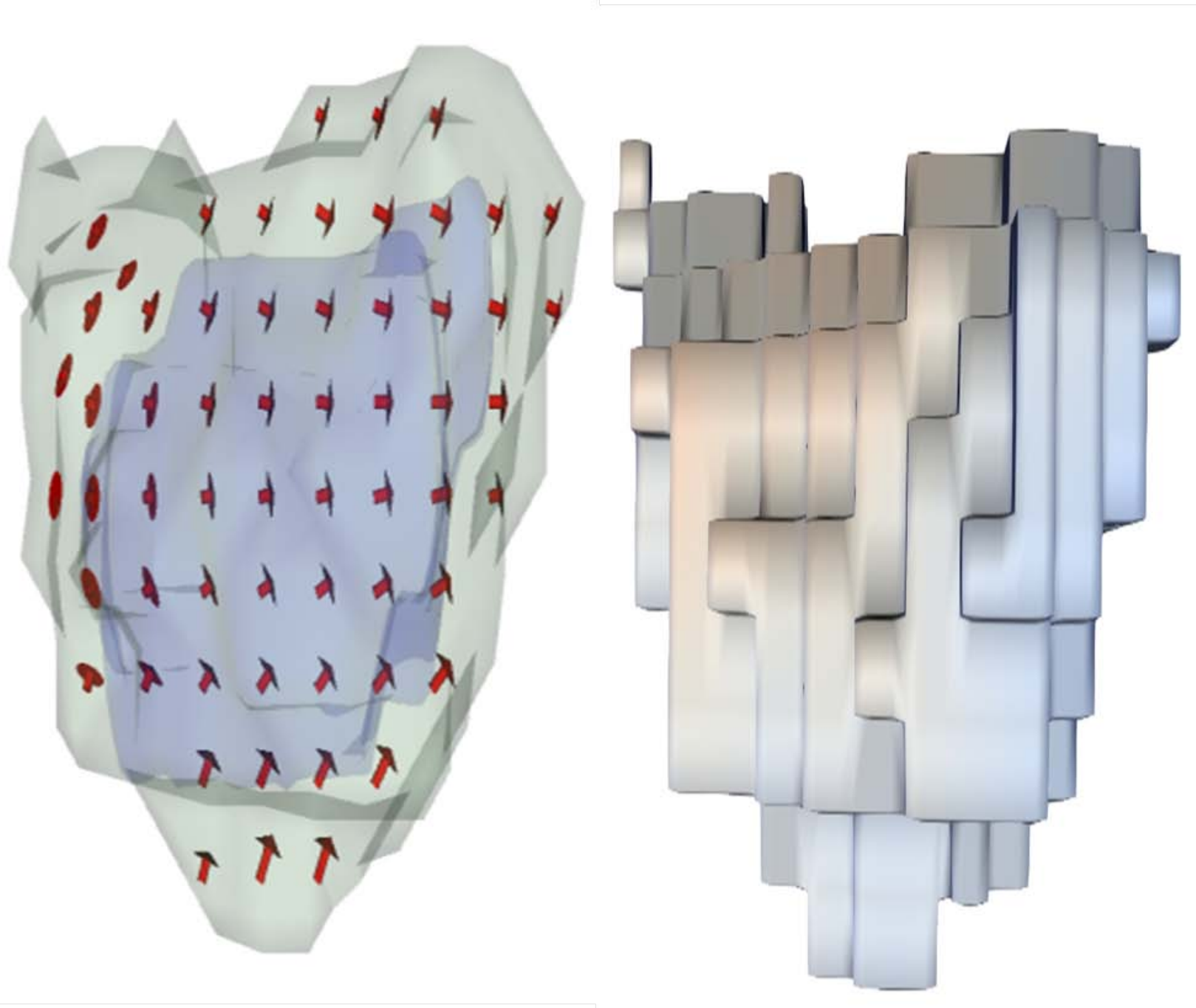}& 
\includegraphics[width=0.03\textwidth]{figures/main_fig/Picture16.pdf}\\
\hline
\includegraphics[width=0.13\textwidth]{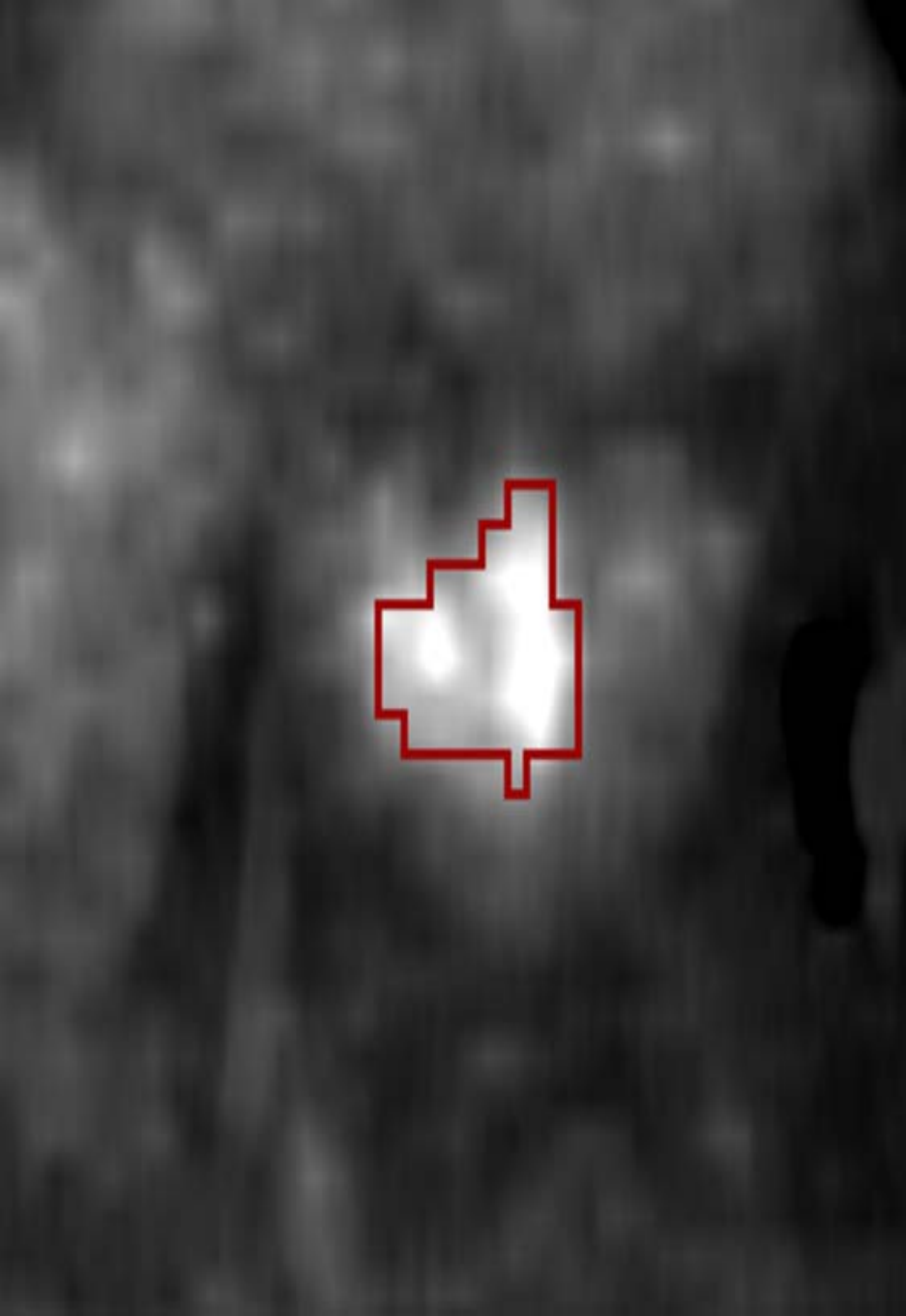}&
\includegraphics[width=0.13\textwidth]{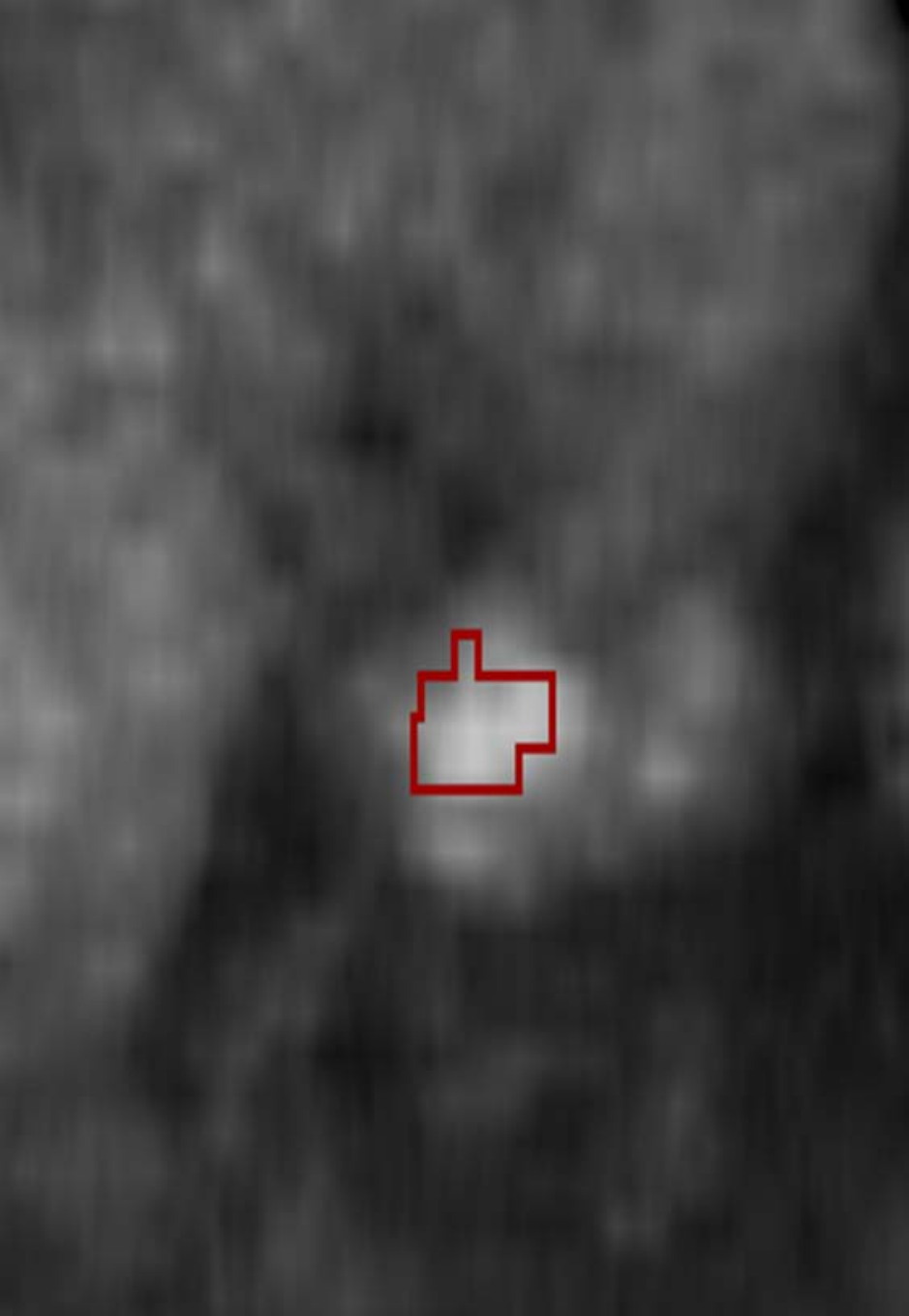}&
\includegraphics[width=0.20\textwidth]{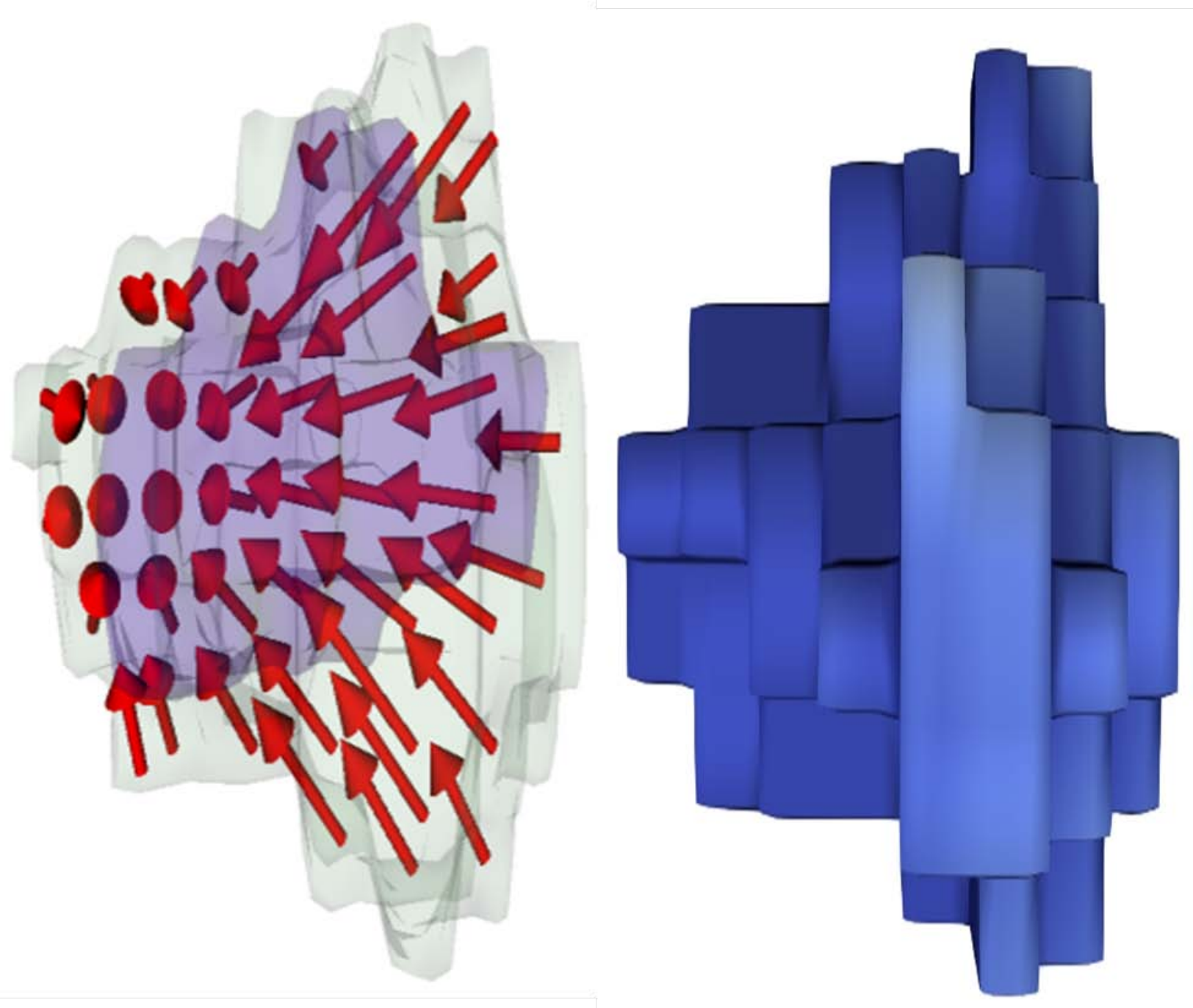}&
\includegraphics[width=0.20\textwidth]{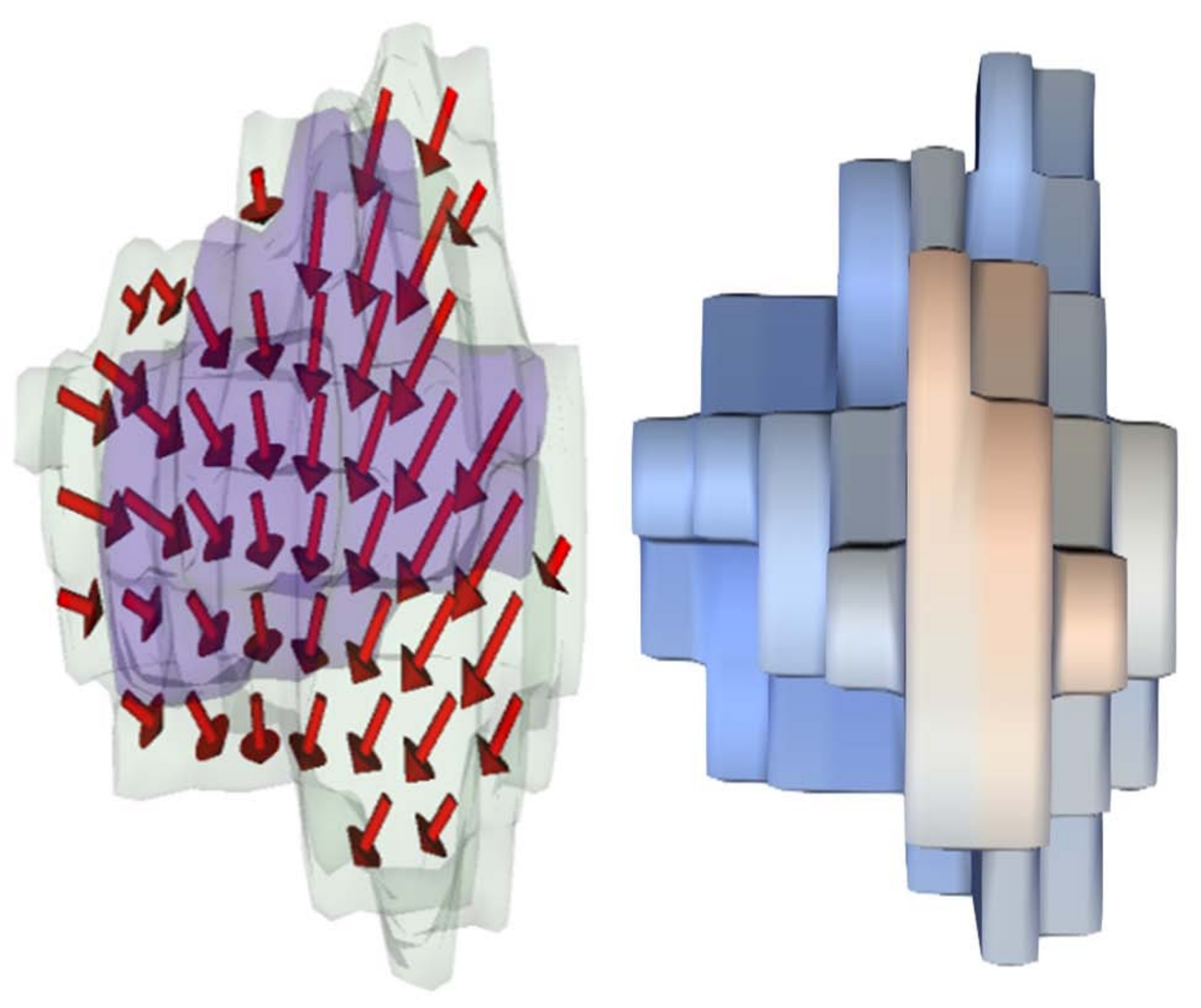}&
\includegraphics[width=0.20\textwidth]{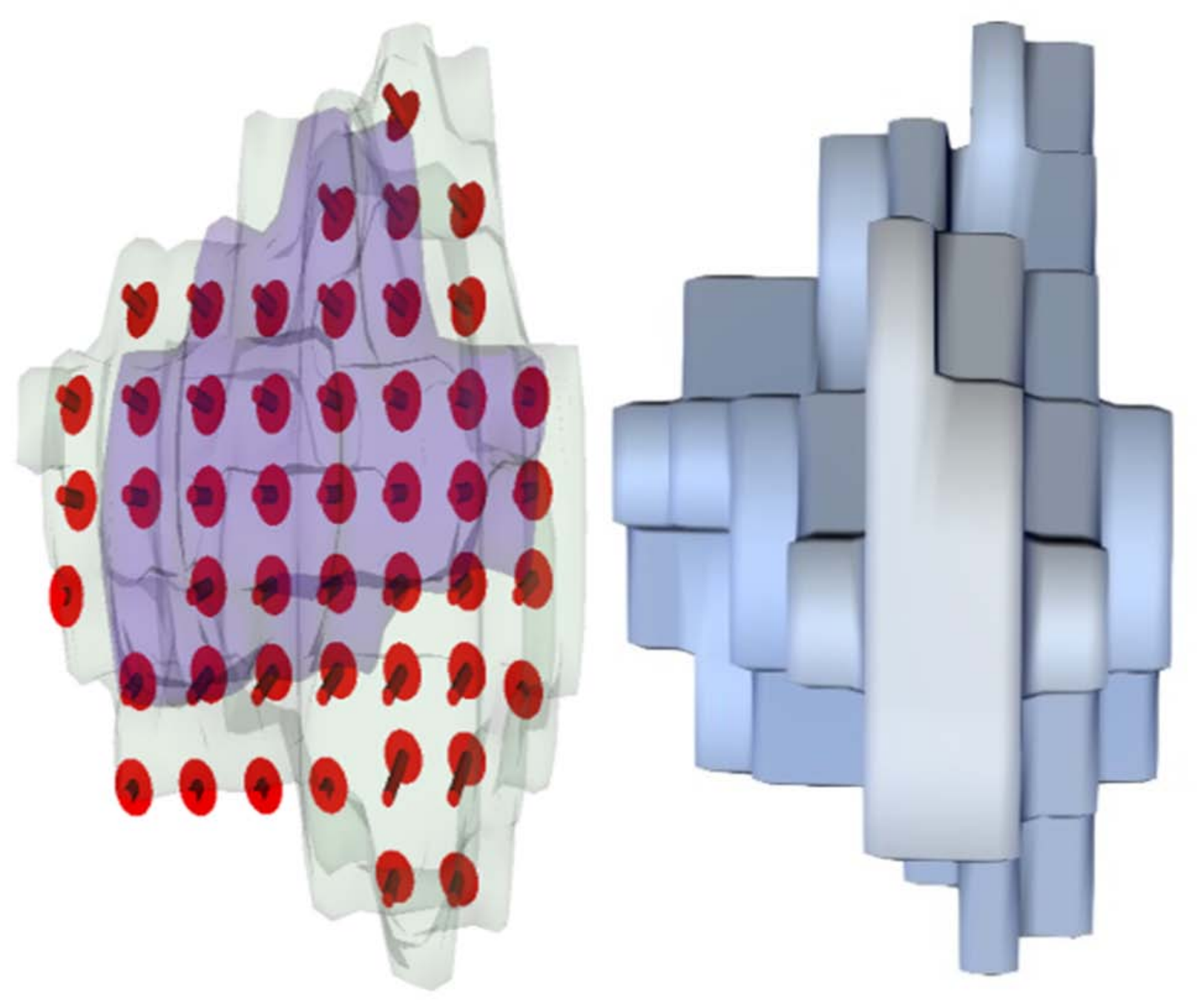}&
\includegraphics[width=0.03\textwidth]{figures/main_fig/Picture16.pdf}\\
\hline
\includegraphics[width=0.13\textwidth]{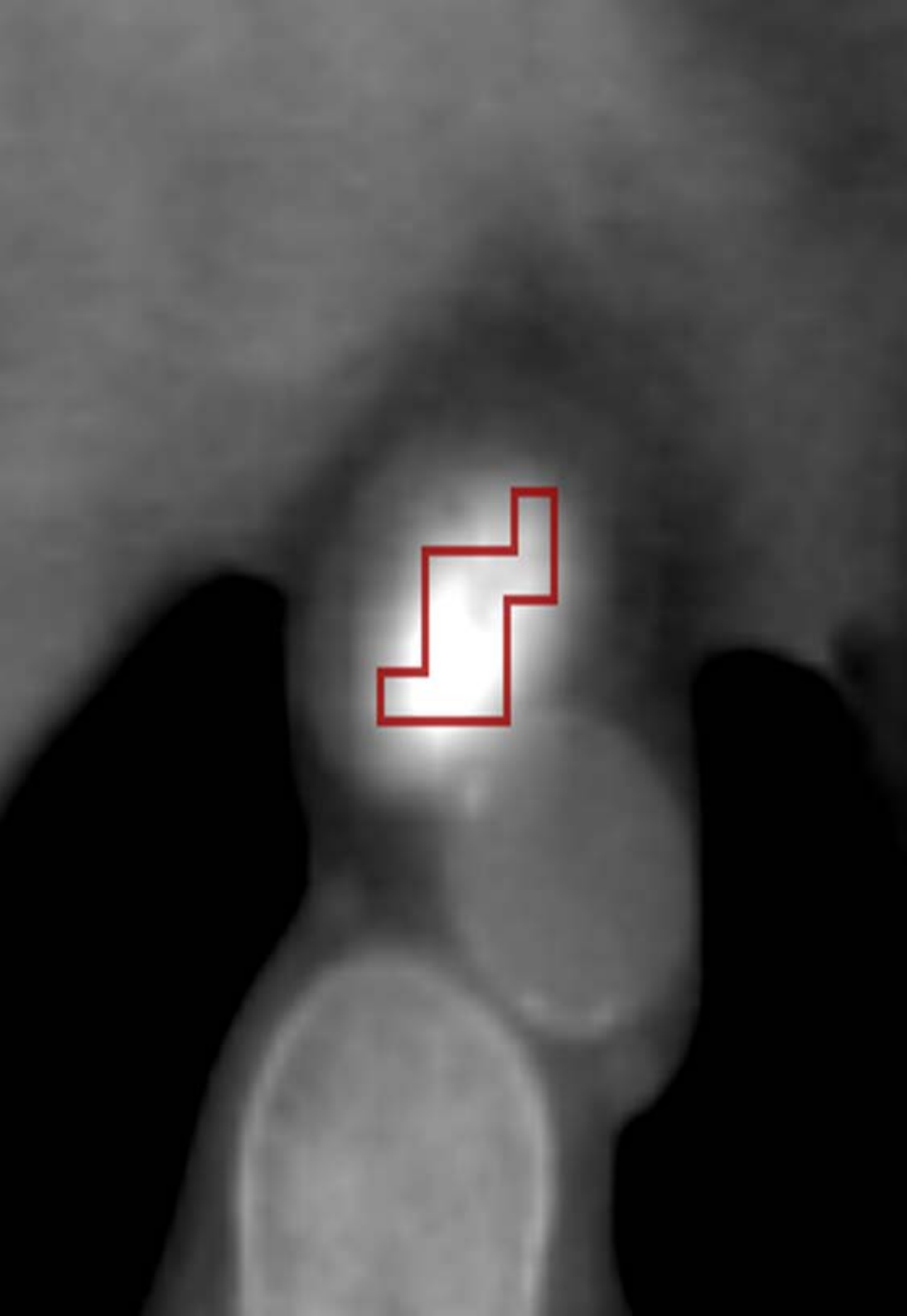}&
\includegraphics[width=0.13\textwidth]{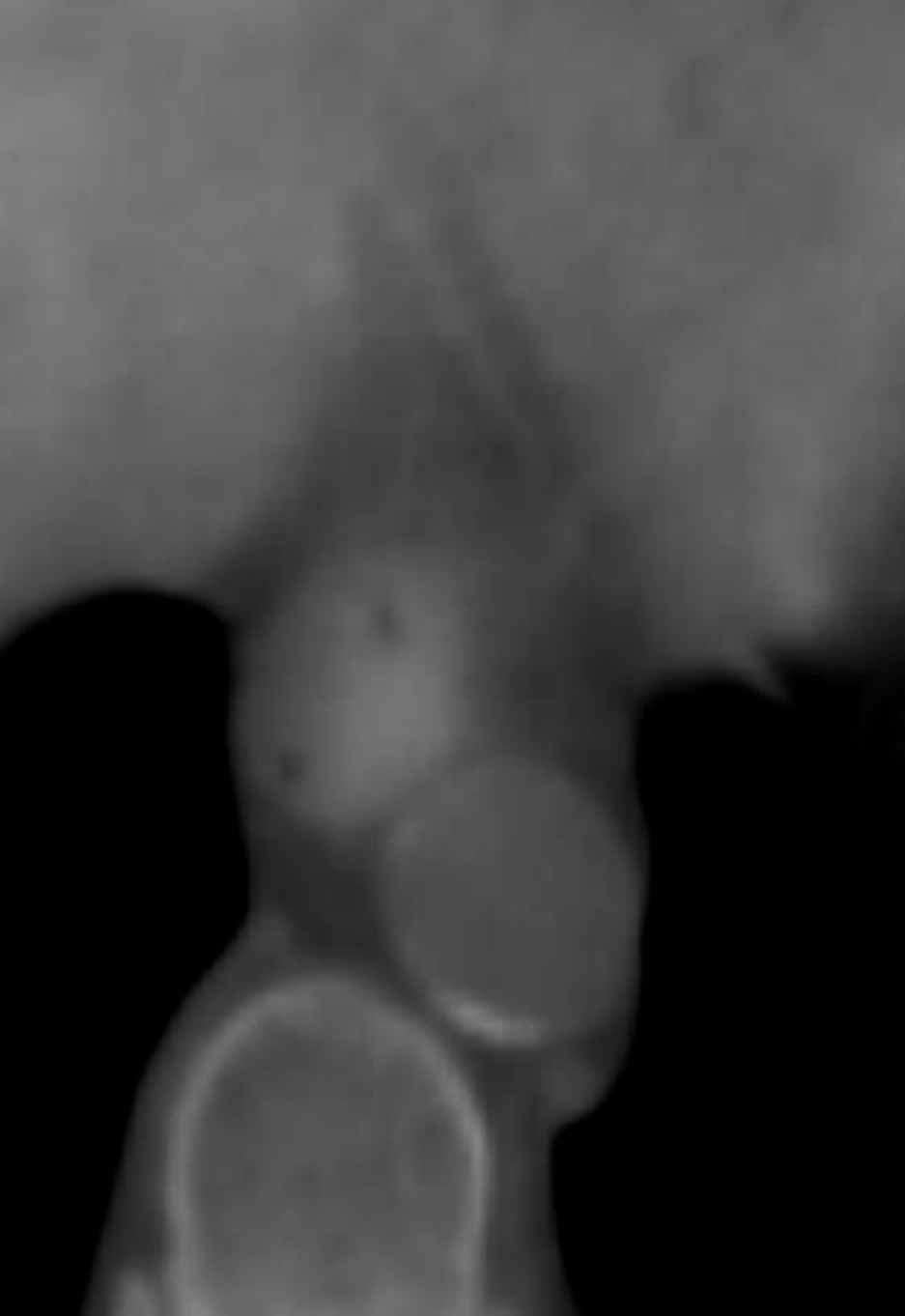}&
\includegraphics[width=0.20\textwidth]{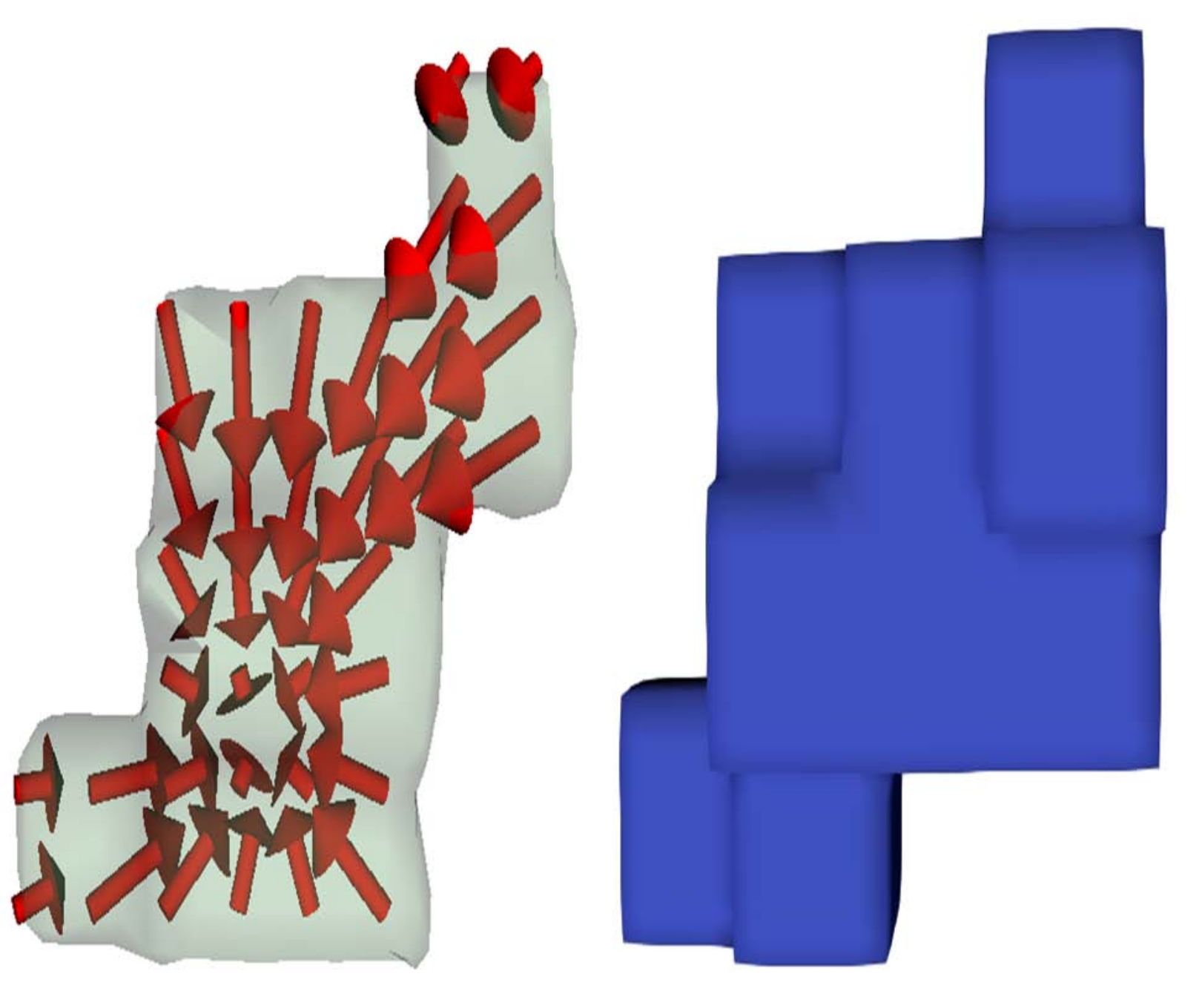}&
\includegraphics[width=0.20\textwidth]{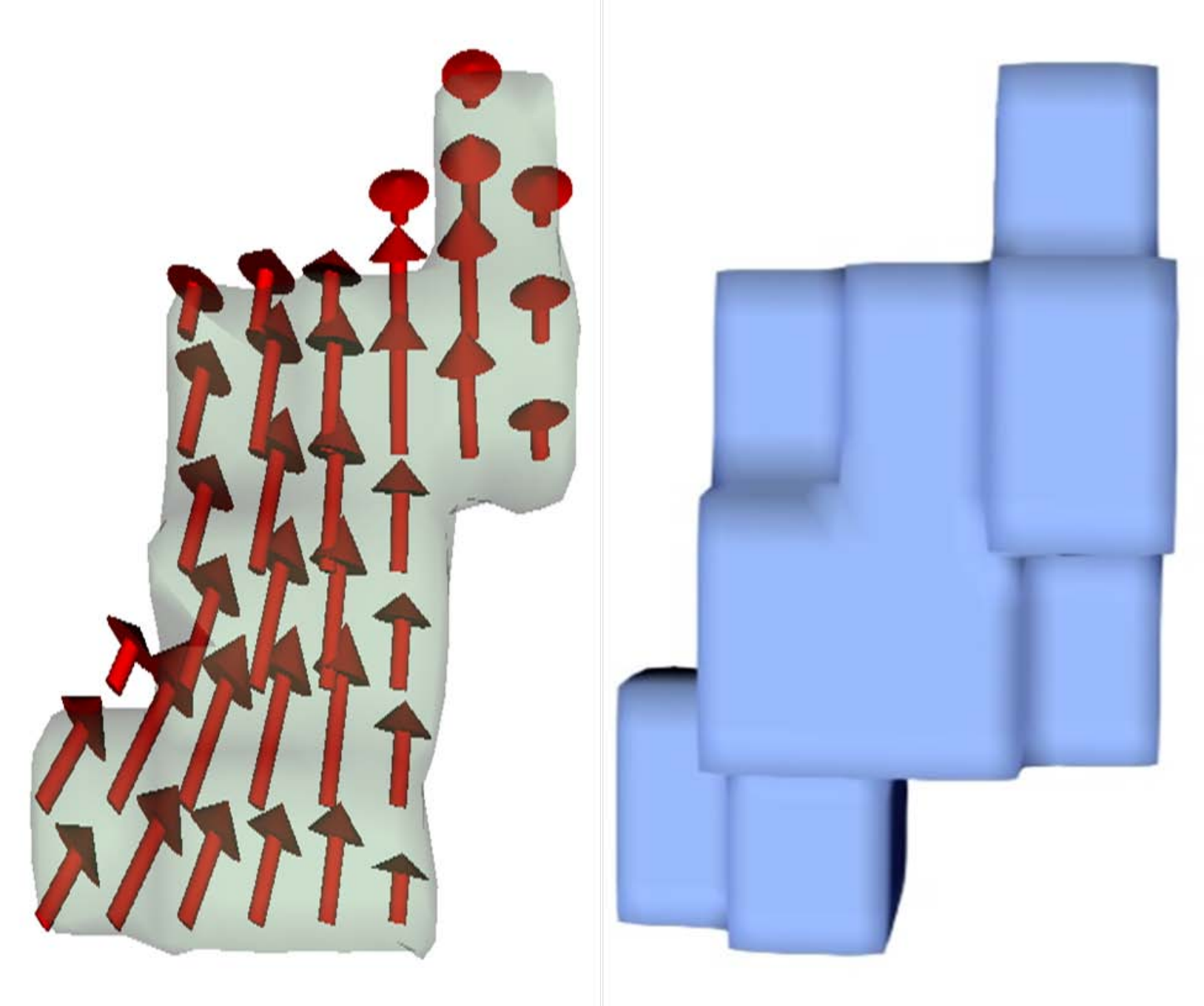}&
\includegraphics[width=0.20\textwidth]{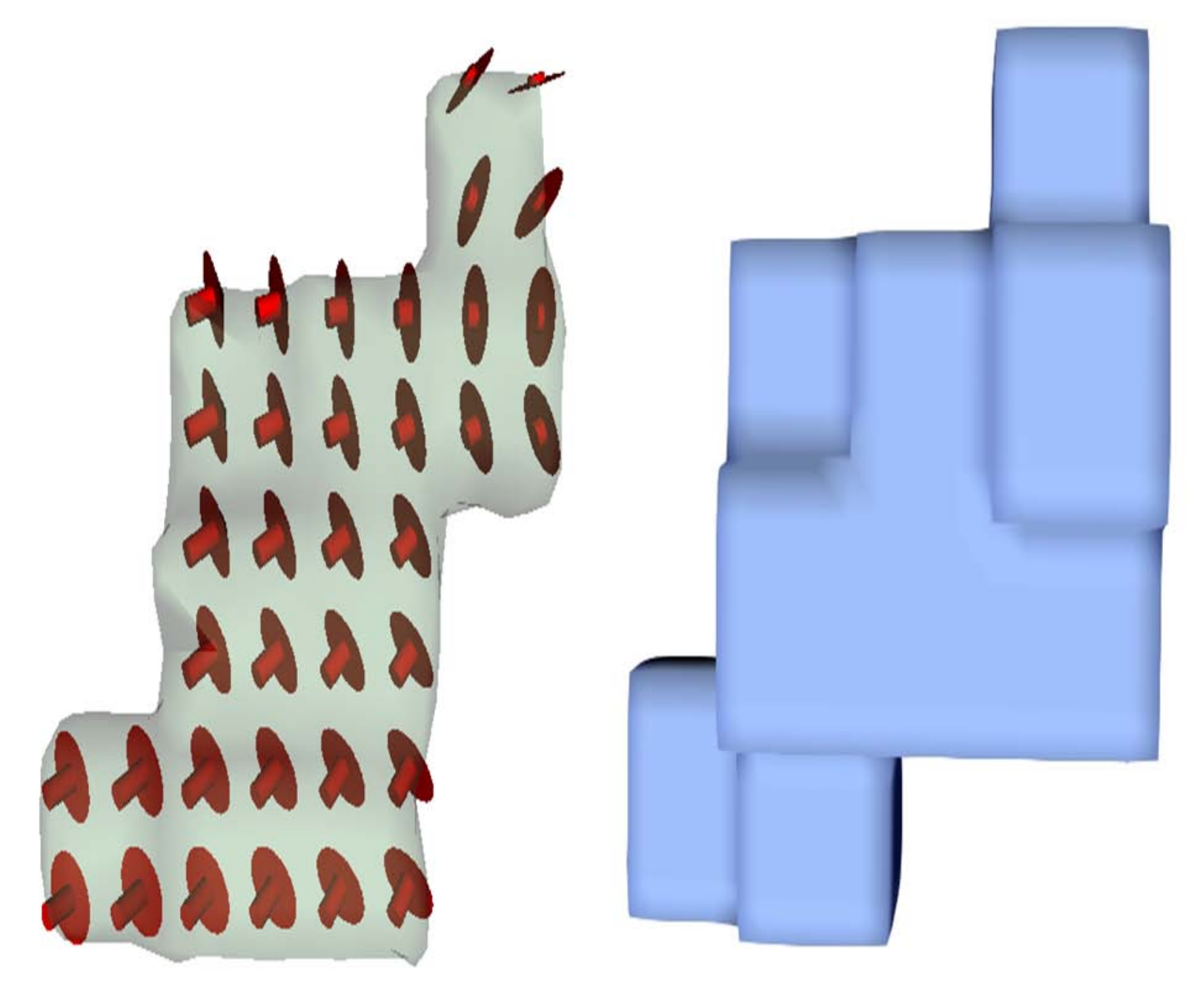}&
\includegraphics[width=0.03\textwidth]{figures/main_fig/Picture16.pdf}\\
\end{tabular}
\end{center}
\caption{First column shows baseline and follow-up blended PET-CT images for three tumors in coronal (top, middle) and axial (bottom) views. Red contour is MTV. In the second to the last column, DVF (left) illustrate the change from baseline MTV (green) to follow-up MTV (blue) and Jacobian maps (right) are overlaid on baseline MTV. Color bar indicates shrinkage (blue) to expansion (red) in Jacobian map.
\label{fig:main_fig}}
\end{figure*}

Fig. \ref{fig:main_fig} shows blended PET-CT images of 3 heterogeneous tumor cases. Tumor shrinkage calculated by blended PET-CT, PET-PET and CT-CT registrations are illustrated using DVF and Jacobian map for each case (Top, Middle and Bottom). Qualitatively, using blended PET-CT image registration, vectors converged from the boundary toward the center of baseline and follow-up MTVs (green and blue volume), generated a sink point in the center where Jacobian was much smaller than 1 (shown in blue in Jacobian map), indicating a large shrinkage. Using PET-PET registration, due to lack of image features, the registration couldn't accurately find the corresponding points and DVF only converged in the tumor boundary. For CT-CT registration, due to smaller structural change and uniform intensity in soft tissue, DVF magnitude was small and Jacobian map mostly showed no volume change. The percentage of tumor shrinkage calculated by semi-automatic segmentation (ground-truth) is listed in Table \ref{table:forfig2} for each case (Top, Middle and Bottom). The percentage of tumor shrinkage calculated by blended PET-CT, PET-PET and CT-CT registrations using both BSD and FFD are also shown in Table \ref{table:forfig2} for each case. Quantitatively, using BSD, both PET-PET and CT-CT registrations showed inferior results compared to blended PET-CT. For smaller shrinkage, FFD had similar accuracy to BSD, but its accuracy worsened for larger changes. However using FFD, PET-PET had the worst results while CT-CT achieved much better accuracy. These results aligned with the literature that diffeomorphic algorithm performs better on larger deformations whereas smaller soft tissue changes in CT can be better captured using the FFD algorithm \cite{ashburner2007fast}.

Jacobian maps in Fig. \ref{fig:main_fig} illustrate local non-uniform tumor changes. Quantifying change in a non-residual tumor (Fig. \ref{fig:main_fig} bottom) using DIR is challenging due to a large non-correspondence deformation between the two images. Here, we showed that using blended PET-CT image registration we could generate a DVF to quite accurately measure tumor change owing to the dominant metabolic tumor structures in the baseline image and anatomical structures in the follow-up image that guided the registration.

\begin{table}[t!]
\centering
\caption{Tumor shrinkage quantified by blended PET-CT, PET-PET and CT-CT registrations compared with ground truth segmentation for each case in Fig. \ref{fig:main_fig}.}
\label{table:forfig2}
\setlength{\tabcolsep}{6.5pt}
\begin{tabular}{c|c|c|c|c|c}
\hline
\bf Registration & \bf Cases & \bf Segmentation & \bf PET-CT & \bf PET-PET & \bf CT-CT\\
\hline
\multirow{3}{*}{BSD} & Top & 51\% & 35\% & 17\% & 17\% \\
& Middle & 78.5\% & 58.5\% & 16\% & 14\% \\
& Bottom & 100\% & 74.5\% & 19\% & 14\% \\
\hline
\multirow{3}{*}{FFD} & Top & 51\% & 35\% & 5.8\% & 14\%\\
& Middle & 78.5\% & 20\% & 2.3\% & 43\%\\
& Bottom & 100\% & 35\% & 6\% & 37\%\\
\hline
\end{tabular}
\end{table}

\subsection{Pathologic Tumor Response Prediction} 
\begin{figure*}[htb]
\begin{center}
\begin{tabular}{ccc}
\includegraphics[width=0.353\textwidth,height=0.335\textwidth]{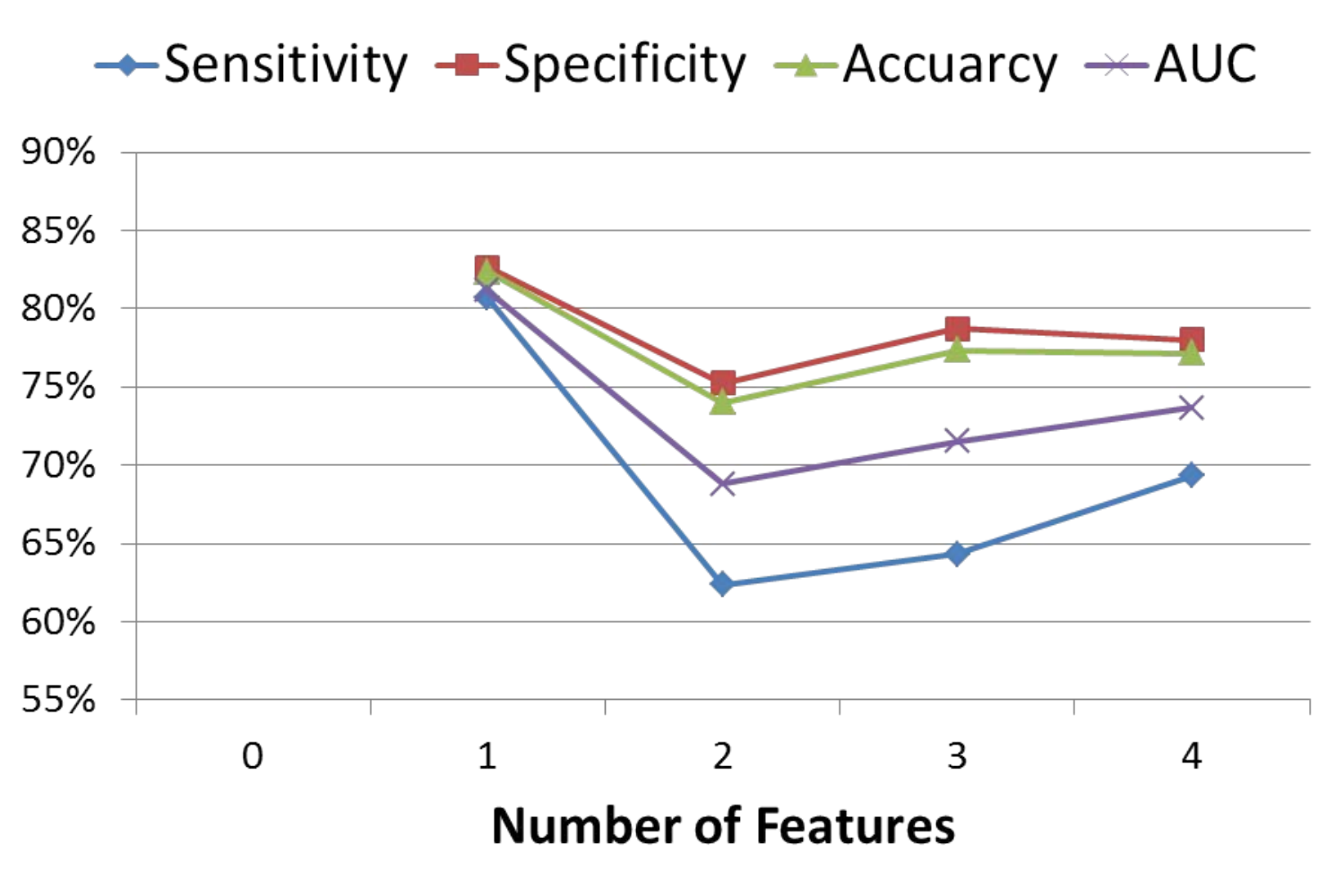}&
\includegraphics[width=0.335\textwidth,height=0.335\textwidth]{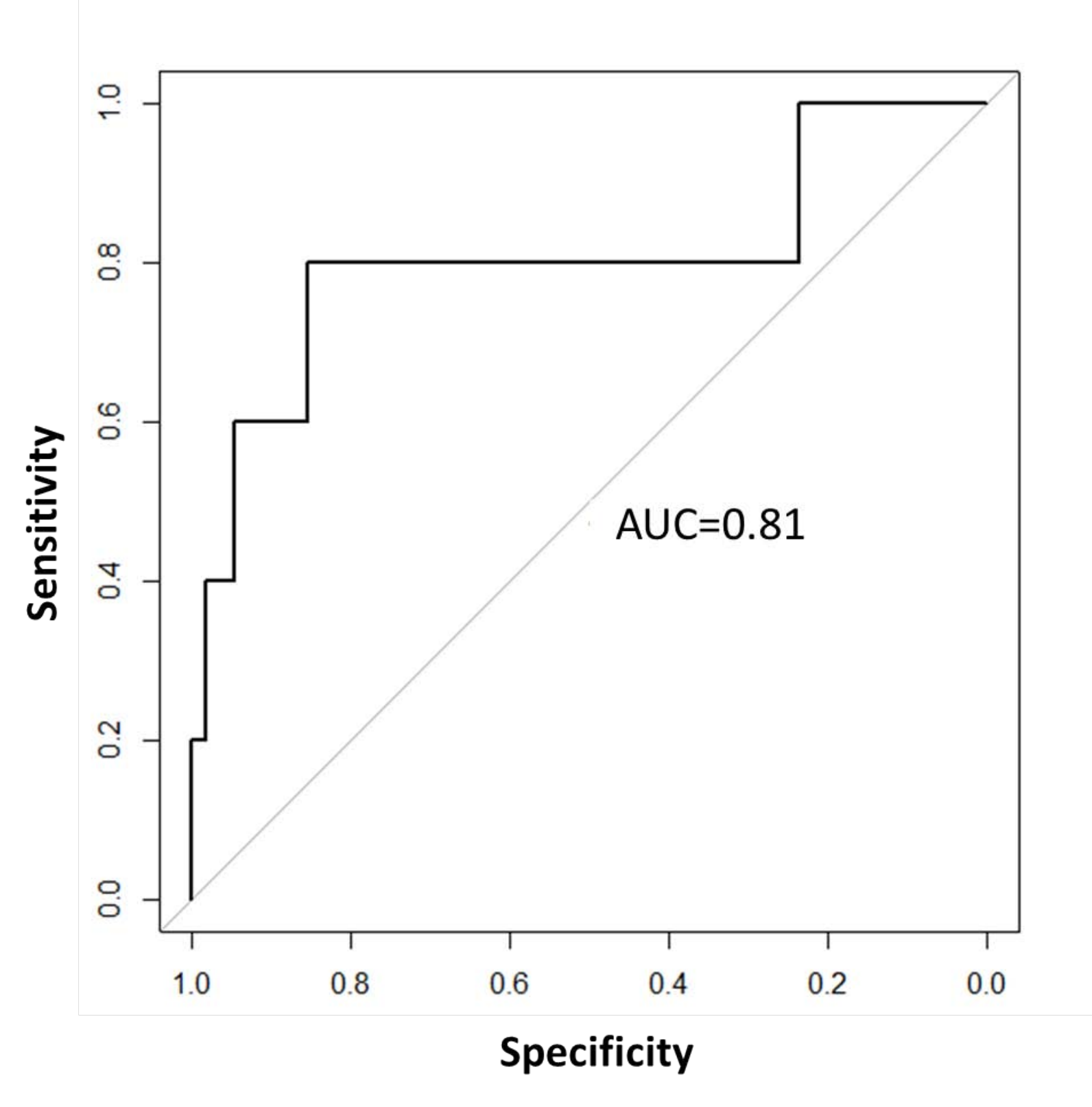}&
\includegraphics[width=0.335\textwidth,height=0.335\textwidth]{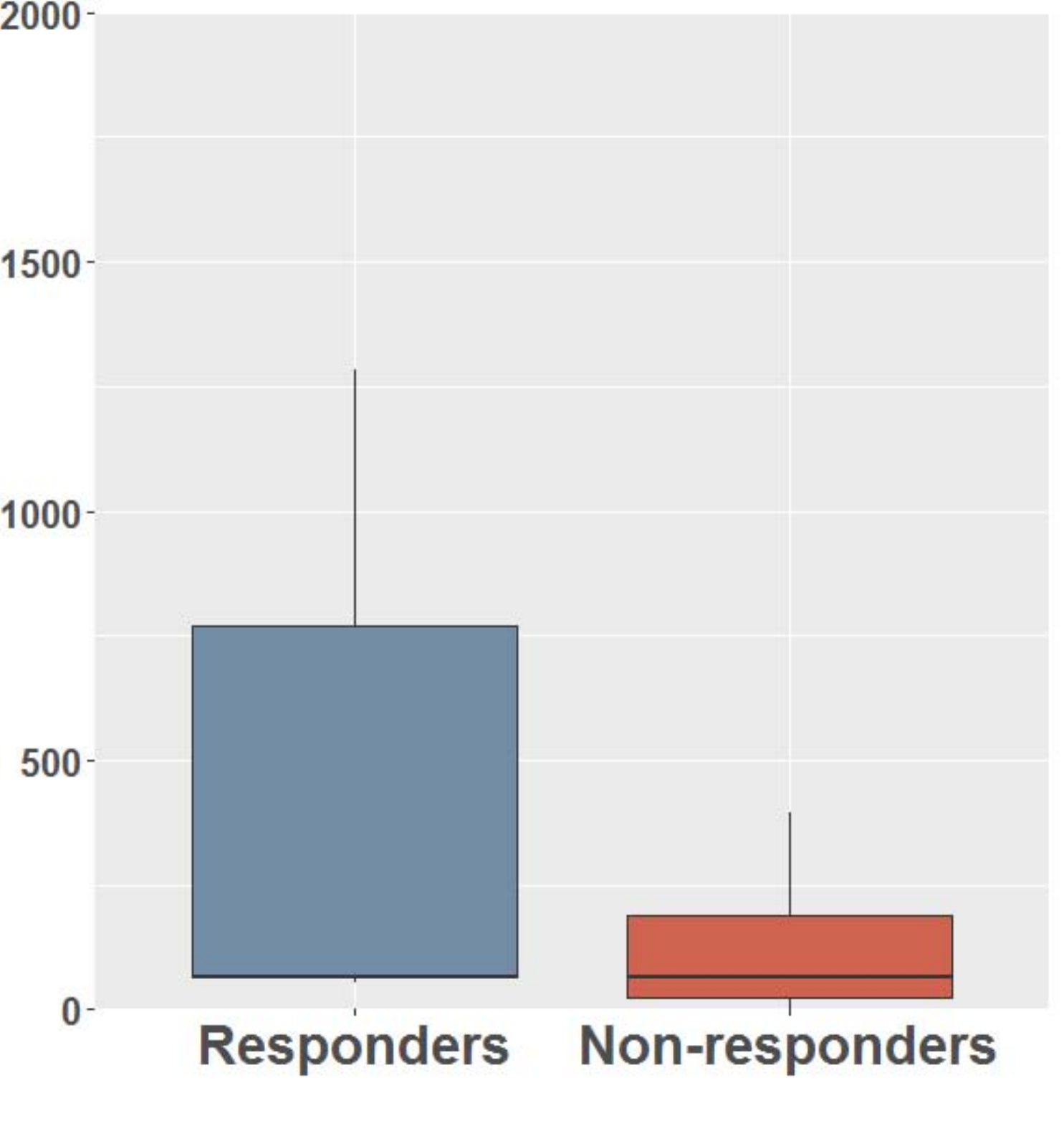}\\
(a) & (b) & (c)
\end{tabular}
\end{center}
\caption{(a)Model performance with increasing number of features (b)ROC curve on the best model (c)Box plot of Mean of Cluster Shade Jacobian feature.
\label{fig:boxplots}}
\end{figure*}

Table \ref{table:uni_analysis} lists the p-value and AUC for all predictive Jacobian features compared with clinical features as well as a recent esophageal cancer radiomics study using univariate analysis. Standard Deviation (SD) of Correlation, a texture feature in Jacobian map of tumors using blended PET-CT BSD registration achieved higher AUC=0.85 compared to PET radiomic features analysis performed by Yip \emph{et al}.~\cite{yip2016textureregistration}. Clinical features in our study were not predictive and none FFD based Jacobian features were significant in differentiating pathologic response.
\begin{table}[t!]
\centering
\caption{Important Jacobian and clinical features in univariate analysis.}
\label{table:uni_analysis}
\setlength{\tabcolsep}{6pt}
\begin{tabular}{l|c|c|c}
\hline
\bf Study & \bf Features & \bf AUC & \bf \textit{p-value} \\
\hline
Yip \emph{et~al.}~\cite{yip2016textureregistration} & Run length matrix	 & 0.71$\sim$0.81 & p$<$0.02\\
\hline
\multirow{2}{*}{Current study} &	$\Delta$MTV	 & 0.62 & 0.33\\
 &	$\Delta$SUV$_{max}$	 & 0.53 & 0.81\\
\hline
\multirow{3}{*}{Blended PET-CT} & SD Correlation & 0.85 & 0.006\\
& SD Energy & 0.80 & 0.01\\
& Mean Cluster Shade & 0.77 & 0.03\\
\hline
\multirow{3}{*}{PET-PET} & Mean Haralick Correlation & 0.81 &	0.01\\
& Mean Entropy & 0.80 &	0.02\\
& Mean Energy & 0.75 &	0.04\\
\hline
\multirow{3}{*}{CT-CT} & SD Long Run High Grey Level & 0.79 & 0.02\\
& SD Long Run & 0.76 & 0.04\\
& SD High Grey Level & 0.76 & 0.04\\
\hline
\end{tabular}
\end{table}
\newline
\noindent
In multivariate analysis, the RF-LASSO model achieved the highest accuracy with only one texture feature - Mean of Cluster Shade extracted from blended PET-CT BSD Jacobian map (Sensitivity=80.6\%, Specificity=82.6\%, Accuracy= 82.3\%, AUC=0.81). However, the performance was worsened when adding more features (Fig. \ref{fig:boxplots} (a)). This feature quantified the heterogeneity of the tumor change and responders showed higher values meaning more heterogeneous local MTV changes. Fig. \ref{fig:boxplots} (b) is the ROC curve of the best model and Fig. \ref{fig:boxplots} (c) shows this feature can differentiate response very well. Mean of Cluster Shade was selected as the first feature by LASSO, however SD correlation with the highest AUC in univariate analysis was selected as the third feature in the multivariate model. This may be because LASSO selects the least correlated features and Mean of Cluster Shade had the smallest mean absolute correlation (r=0.22) among the important distinctive features compared to SD correlation (r=0.46).

\section{Conclusion and Future Work}
We combined PET and CT images into a grayscale blended PET-CT image for quantification of local metabolic tumor change using Jacobian map. We extracted intensity and texture features from the Jacobian map to predict pathologic tumor response in esophageal cancer patients. Jacobian texture features showed the highest accuracy for prediction of pathologic tumor response (accuracy=82.3\%). In the future, we will explore automated optimal weight tuning for PET-CT blending.


\bibliographystyle{splncs03}

\end{document}